%% file: VisEvol.tex
\documentclass{egpubl}
\usepackage{eurovis2021}

\usepackage[absolute,overlay]{textpos}
\usepackage{pbox}
\usepackage{everypage}

\newcommand{\stamp}[1][© 2021 The Eurographics Association and John Wiley \& Sons Ltd. This is the author's version of the article that has been published in Computer Graphics Forum. The final version of this record is available at: \href{https://doi.org/10.1111/cgf.14300}{\color{blue}10.1111/cgf.14300}]{%
\begin{textblock*}{140mm}(37mm,270mm)
\centering%
\small%
\emph{#1}%
\end{textblock*}%
}

\AddEverypageHook{
  \stamp
}

\SpecialIssuePaper         
\usepackage[T1]{fontenc}
\usepackage{dfadobe}  

\usepackage[normalem]{ulem}

\usepackage[dvipsnames]{xcolor} 
\newcommand{\hl}[1]{#1} 
\newcommand{\circled}[1]{\raisebox{.4pt}{\textcircled{\raisebox{0.15pt}{\resizebox{!}{0.8ex}{\textbf{\textsf{#1}}}}}}}

\usepackage{cite}  
\BibtexOrBiblatex
\electronicVersion
\PrintedOrElectronic
\ifpdf \usepackage[pdftex]{graphicx} \pdfcompresslevel=9
\else \usepackage[dvips]{graphicx} \fi
\graphicspath{{figures/}{pictures/}{images/}{./}} 

\usepackage{cleveref}
\renewcommand{\autoref}{\Cref}

\usepackage{egweblnk}

\usepackage{longfbox}

\newfboxstyle{tight2}{padding=2pt,margin=0pt,baseline-skip=false}%

\definecolor{ColT1}{HTML}{1f77b4}
\definecolor{ColT2}{HTML}{ff7f0e}
\definecolor{ColT3}{HTML}{2ca02c}
\definecolor{ColT4}{HTML}{d62728}
\definecolor{ColT5}{HTML}{9467bd}
\definecolor{ColT6}{HTML}{8c564b}
\definecolor{ColT7}{HTML}{e377c2}
\definecolor{ColT8}{HTML}{7f7f7f}
\definecolor{ColT9}{HTML}{bcbd22}
\definecolor{ColT10}{HTML}{17becf}

\newlength{\boxh}
\title[VisEvol: Visual Analytics to Support Hyperparameter Search through Evolutionary Optimization]%
      {VisEvol: Visual Analytics to Support Hyperparameter Search through Evolutionary Optimization}

\author[Chatzimparmpas et al.]
{\parbox{\textwidth}{\centering \vspace{-1.7cm} A. Chatzimparmpas$^{1}$\orcid{0000-0002-9079-2376}, R.\,M. Martins$^{1}$\orcid{0000-0002-2901-935X}, K. Kucher$^{1}$\orcid{0000-0002-1907-7820}, and A. Kerren$^{1,2}$\orcid{0000-0002-0519-2537}}
        \\
{\parbox{\textwidth}{\centering \vspace{-1.8cm} $^1$Department of Computer Science and Media Technology, Linnaeus University, Sweden\\
$^2$Department of Science and Technology, Linköping University, Sweden}}
}

\teaser{
 \centering
 \vspace{-1.9cm}
 \includegraphics[width=\linewidth]{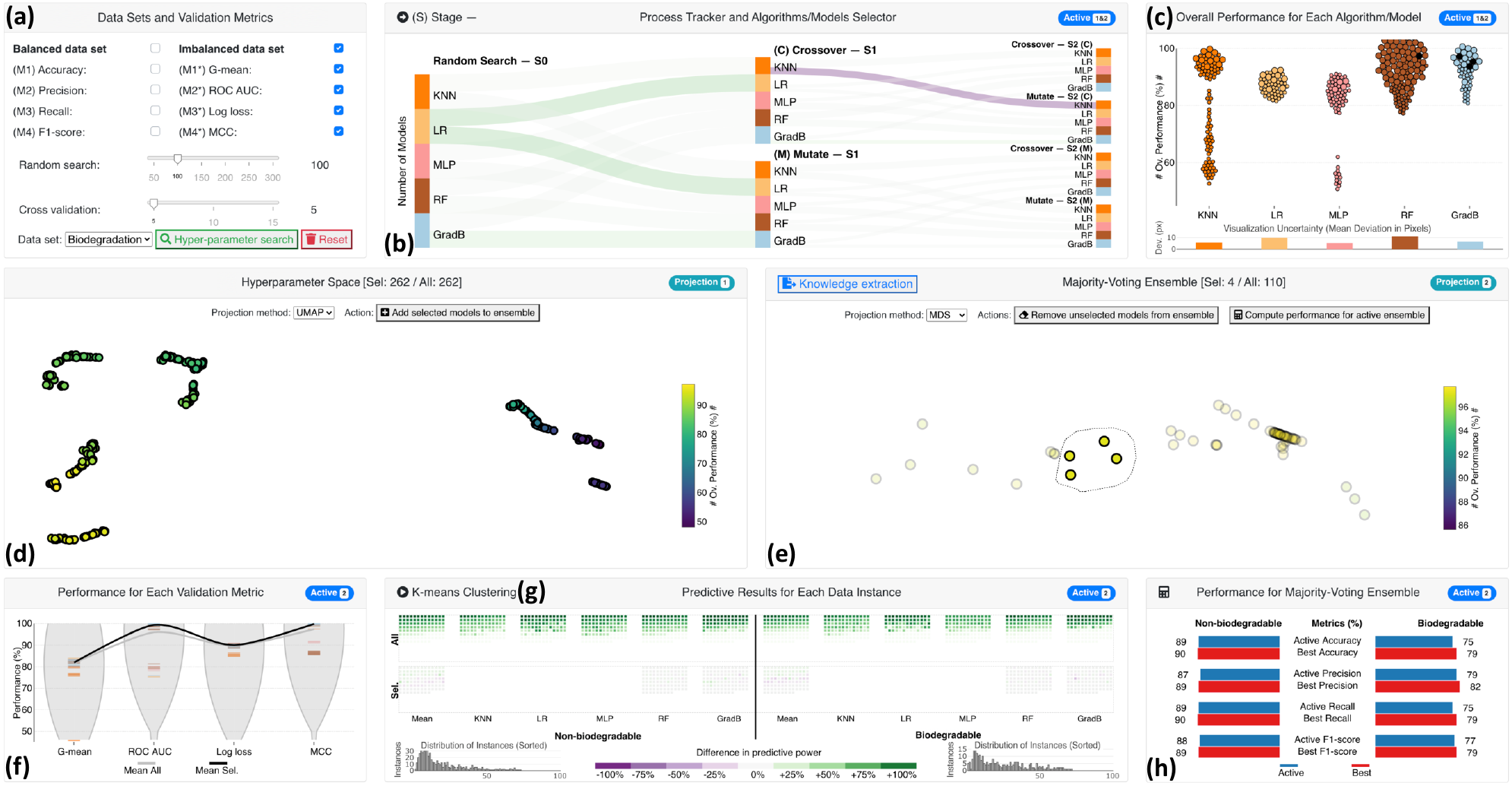}
 \caption{Controlling the evolutionary optimization process for hyperparameter search with VisEvol: (a) the panel for the selection of validation metrics and initialization of random search; (b) the Sankey diagram for the management and analysis of the crossover and mutation procedure; (c) the beeswarm plot with sorted algorithms/models according to overall performance; (d) the projection-based visualization of hyperparameters that aggregates the results of the chosen metrics for all models; (e) the visual embedding of ensembles that includes the handpicked models; (f) the bean plot that presents the performance of models for each metric; (g) the grid-based visualization that displays the predictive power in each instance; and (h) the horizontal bar chart for showing the results of the active vs. the best voting ensemble.
 }
	\label{fig:teaser}
}
\begin{document}

\stamp

\maketitle

\begin{abstract}
   \input{0.Abstract}

   \makeatletter

	\def\customclassification{\vskip 5.5pt\par\reset@font\rmfamily}
	\def\endcustomclassification{\relax}
	\makeatother

	\begin{customclassification}
		\textbf{CCS Concepts}\\
		$\bullet$ \textbf{Human-centered computing} $\rightarrow$ Visualization; Visual analytics; 
		$\bullet$ \textbf{Machine learning} $\rightarrow$ Supervised learning;
	\end{customclassification}
\end{abstract} 

\section{Introduction} \label{sec:intro}
	\input{1.Introduction}

\section{Related Work} \label{sec:relwo}
	\input{2.Related_work}

\section{Analytical Requirements and Design Goals} \label{sec:goals}
	\input{3.Design_goals}

\section{VisEvol: System Overview and Application} \label{sec:overview}
	\input{4.System_overview}

\section{Use Case} \label{sec:case}
	\input{5.Use_case}

\section{Evaluation} \label{sec:eval}
	\input{6.Evaluation}

\section{Conclusion} \label{sec:con}
	\input{7.Conclusion}

\bibliographystyle{eg-alpha-doi}

\bibliography{references}

\end{document}

%% file: 0.Abstract.tex
During the training phase of machine learning (ML) models, it is usually necessary to configure several hyperparameters.  This process is computationally intensive and requires an extensive search to infer the best hyperparameter set for the given problem. The challenge is exacerbated by the fact that most ML models are complex internally, and training involves trial-and-error processes that could remarkably affect the predictive result. 
Moreover, each hyperparameter of an ML algorithm is potentially intertwined with the others, and changing it might result in unforeseeable impacts on the remaining hyperparameters.
Evolutionary optimization is a promising method to try and address those issues. According to this method, performant models are stored, while the remainder are improved through crossover and mutation processes inspired by genetic algorithms. We present VisEvol, a visual analytics tool that supports interactive exploration of hyperparameters and intervention in this evolutionary procedure. In summary, our proposed tool helps the user to generate new models through evolution and eventually explore powerful hyperparameter combinations in diverse regions of the extensive hyperparameter space. The outcome is a voting ensemble (with equal rights) that boosts the final predictive performance. The utility and applicability of VisEvol are demonstrated with two use cases
and interviews with ML experts who evaluated the effectiveness of the tool.

%% file: 1.Introduction.tex
Hyperparameter optimization (also called \emph{hyperparameter tuning}) is the process of selecting appropriate values of hyperparameters for machine learning (ML) models, often independently for each data set, to achieve their best possible results.
Although time consuming, this process is required for the vast majority of ML models before their deployment into production~\cite{Rijn2017An,Rijn2018Hyperparameter}. 
%
Numerous techniques exist that try to solve this challenge, such as the well-known \emph{grid search}, \emph{random search}~\cite{Bergstra2012Random}, and \emph{Bayesian optimization} that belong to the generic type of \emph{sequential-based methods}~\cite{Bergstra2011Algorithms,Shahriari2016Taking}. Other proposed methods include \emph{bandit-based approaches}~\cite{Falkner2018BOHB,Li2017Hyperband}, \emph{population-based methods}~\cite{Jaderberg2017Population}, and \emph{evolutionary optimization}~\cite{Francescomarino2018Genetic,Young2015Optimizing}, which is our focus in this paper. 

\hl{Inspired by the biological concept of evolution, one strategy of \emph{evolutionary optimization} keeps the top-ranked models and replaces the remaining, worst-performing models using new hyperparameter sets generated through \emph{crossover} and \emph{mutation}~\cite{Cantu2005An}.}
With \emph{crossover}, random pairs of underperforming models (originating from the same algorithm) are picked and their hyperparame\-ters are fused with the goal of creating a better model. As a result, internal regions of the solution space are further explored, and better local optima are investigated. 
On the other hand, \emph{mutation} randomly generates new values for the hyperparameters to substitute old values. It facilitates scanning for external regions of the solution space to discover additional local optima. These unexplored areas of the hyperparameter space may offer a fresh start to the search for hyperparameters. The synergy of combining both techniques can be beneficial in finding distinctive local optima that generalize to a better result in the end. Hence, the problem of getting stuck in local optima of the hyperparameter space is addressed. However, one question that emerges is: \textbf{(RQ1)} how to choose which models (and algorithms) should crossover and/or mutate, and to what extent, considering we have limited computational resources?

Various automatic ML methods~\cite{Feurer2019Hyperparameter} and practical frameworks~\cite{comet,nni} have been proposed to deal with the challenge of hyperparameter search. However, their output is usually a single model, which is frequently underpowered when compared to an ensemble of ML models~\cite{Sagi2018Ensemble}. 
Ensemble methods---such as \emph{bagging} and \emph{boosting}---could be combined in a majority-voting ensemble~\cite{Cai2013A} with a democratic voting system that summarizes the decisions among models.
The authors of a recent survey~\cite{Sagi2018Ensemble} state that users should understand how to tune models and, in extension, choose hyperparameters for selecting the appropriate ML ensemble. Consequently, another open question is: \textbf{(RQ2)} how to find which particular hyperparameter set is suitable for each model in a majority-voting ensemble of diverse models?

The optimization of hyperparameters is often performed with the support of a single, specific validation metric (e.g., in Bayesian optimization)~\cite{Snoek2012Practical}.
The selection of a proper metric for a task is related to the particular data set, the problem, and the tasks at hand. 
Thus, the use of a single metric for every data set (such as, e.g., accuracy~\cite{McNee2006Being,Sturm2013Classification}) may result in several problems.
The use of multiple metrics, however, poses an extra challenge to such an automatic optimization procedure~\cite{Ferri2009An,Pereira2017A,Sokolova2009Performance,Tharwat2018Classification}, and the comparison and selection between multiple performance indicators are not trivial, even for widely used metrics~\cite{Davis2006The,Saito2015The}. 
Alternatives, such as Matthews correlation coefficient (MCC), might be more informative for imbalanced classification data~\cite{Chicco2020The}, but
even advanced metrics are not the holy grail, and additional challenges can be found in the literature~\cite{Lobo2008AUC,Powers2011Evaluation}.
This leads to one further question: \textbf{(RQ3)} is there any performance improvement from employing several validation metrics that fit better to a specific data set's inherent characteristics?

Evolutionary optimization and majority-voting ensembles inspired us to focus on the three aforementioned questions that constitute open research challenges. In this paper, we present a visual analytics (VA) tool, called \emph{VisEvol} (see \autoref{fig:teaser}), that addresses the three research questions described above by supporting the exploratory combination of five different ML algorithms. \hl{VisEvol uses validation metrics for balanced and imbalanced data sets, and it involves an initial stage of random search ($S_0$) and two evolutionary generation stages ($S_1$ and $S_2$) of hyperparameter settings (see the details in \autoref{sec:overview}).}
To address the three research questions (\textbf{RQ1--RQ3}), VisEvol supports the following \textbf{workflow} (cf. \autoref{fig:workflow-diagram} described in \autoref{sec:overview}): (i)~the selection and combination of appropriate validation metrics, (ii)~the overall exploration of different algorithms and models using diverse hyperparameters, (iii)~the inspection of predictive impact for each data instance, (iv)~the control of the evolutionary process, and (v)~a final phase where the performance of the current best ML ensemble is compared to the currently active majority-voting ensemble.
In summary, our contributions consist of the following:

\begin{itemize}
\item the systematization of hyperparameter search using evolutionary optimization with a coherent visual analytic workflow;
\item an implementation of the aforementioned conceptual proposal, our VA tool called VisEvol, that consists of a novel combination of interactive coordinated views---which control the crossover and mutation processes---and supports the visual exploration of the most performant/diverse models for the creation of a powerful majority-voting ensemble;
\item a demonstration of the applicability of our proposed system with two use cases, using real-world data, that confirm the effectiveness of controlling the process of evolutionary optimization of hyperparameters and testing different ML ensembles; and
\item the discussion of the methodology of the interviews and the positive and supportive feedback received from three ML experts.
\end{itemize}      

\noindent The rest of this paper is organized as follows. In~\autoref{sec:relwo}, we discuss relevant techniques for visualizing hyperparameter tuning and existent automatic approaches.
Afterwards, in~\autoref{sec:goals}, we describe the analytical requirements and design goals for attaching VA to evolutionary optimization and combining VA with ensemble learning.
\autoref{sec:overview} presents the functionalities of the tool and, at the same time, describes the first use case with the goal of selecting a composition of models (with specific hyperparameters) for the creation of a majority-voting ensemble using medical data.
Thereafter in~\autoref{sec:case}, we demonstrate the applicability and usefulness of Vis\-Evol with another real-world data set focusing on biodegradation of molecules. Next, in~\autoref{sec:eval}, we review the feedback our VA tool obtained during the interview sessions by summing up the experts' opinions and the limitations that guide us to possible future directions for VisEvol. Finally,~\autoref{sec:con} concludes our paper.

%% file: 2.Related_work.tex
Visualization tools have been implemented for sequential-based, bandit-based, and population-based approaches~\cite{Park2021HyperTendril}, and for more straightforward techniques such as grid and random search~\cite{Li2018HyperTuner}. \hl{Evolutionary optimization, however, has not experienced similar consideration by the \mbox{InfoVis} and VA communities, with the exception of more general visualization approaches such as EAVis~\cite{Kerren2005EAVis,Kerren2006Improving} and interactive evolutionary computation (IEC)~\cite{Takagi2001Interactive}.} To the best of our knowledge, there is no literature describing the use of VA in hyperparameter tuning of evolutionary optimization (as defined in \autoref{sec:intro}) with the improvement of performance based on majority-voting ensembles. 
In this section, we review prior work on automatic approaches, visual hyperparameter search, and tools with which users may tune ML ensembles. Finally, we discuss the differences of such systems when compared to VisEvol in order to clarify the novelty of our tool.

\paragraph*{Automatic Approaches.} \label{sec:autom}
In the ML community, most of the research is geared towards fully-automated hyperparameter search with no human interaction~\cite{Claesen2014Easy}. 
It is true that automatic techniques present encouraging results and are successful on tuning hyperparameters of some models~\cite{Bardenet2013Collaborative,Yogatama2014Efficient}, for example, by automatically finding optimal deep learning hyperparameters using genetic algorithms~\cite{Fiszelew2007Finding,Young2015Optimizing}.
Important contributions of this research include the formalization of primary concepts~\cite{Claesen2015Hyperparameter}, the identification of methods for assessing hyperparameter importance~\cite{Jia2016QIM,Probst2019Tunability,Rijn2017An,Hutter2013An,Hutter2014An,Rijn2018Hyperparameter}, and resulting libraries and frameworks for specific hyperparameter optimization methods~\cite{Koch2018Autotune,Thornton2013Auto}. Indeed, several packages exist that focus on automatically optimizing Bayesian methods with the use of a single performance measurement~\cite{BayesOpt,Hutter2011Sequential,Hutter2009ParamILS,Shahriari2016Taking}, and there are popular commercial platforms developed for hyperparameter optimization~\cite{datarobot,automl}. This widespread automation does not stop in supervised classification problems, but also includes dimensionality reduction (DR) algorithms (e.g., t-SNE)~\cite{Belkina2019Automated,Kobak2019The}. 

\hl{Despite the success of automatic approaches and their advancement through the years, it is important to note that such approaches require extensive computing power and may lack critical features.} Automatically (or manually) set thresholds may discard different models which could be informative but theoretically seem to perform worse than the rest.
Moreover, the ranking of models is often based on a single validation metric, leading to the risks discussed in~\autoref{sec:intro}.
The aforementioned works that make use of genetic algorithms contain similar mechanisms as in VisEvol, but without VA support for (1) the exploration of the interconnected hyperparameters, and (2) the selection of the proper number of models that should crossover and mutate. 

\paragraph*{Visual Hyperparameter Searching.} \label{sec:search}
ATMSeer~\cite{Wang2019ATMSeer} implements a multi-granularity visualization for model selection and hyperparameter tuning. It is a visualization tool coupled with a backend framework, called ATM~\cite{Swearingen2017ATM}, that allows the users to interact with the middle steps of an AutoML process and control them by adjusting the search space dynamically during execution time. In contrast to VisEvol, it only supports a single performance measurement, and the output is a single optimized model. 

One common focus of related work is the hyperparameter search for deep learning models. HyperTuner~\cite{Li2018HyperTuner} is an interactive VA system that enables hyperparameter search by using a multi-class confusion matrix for summarizing the predictions and setting user-defined ranges for multiple validation metrics to filter out and evaluate the hyperparameters. Users can intervene in the running procedure to anchor a few hyperparameters and modify others. However, this could be hard to generalize for more than one algorithm at the same time. In our case, we combine the power of diverse algorithms, with one of them being a neural network (NN). HyperTendril~\cite{Park2021HyperTendril} is a visualization tool that supports random search, population-based training~\cite{Jaderberg2017Population}, Bayesian optimization, HyperBand~\cite{Li2017Hyperband}, and the last two methods joined together~\cite{Falkner2018BOHB}. It enables the users to set an initial budget, search the space for the best configuration, and select suitable algorithms. However, its effectiveness is only tested in scenarios specifically designed for NNs.
Other examples of publications which work explicitly with deep learning only, and do not support evolutionary optimization, are VisualHyperTuner~\cite{Park2019VisualHyperTuner}, Jönsson et al.~\cite{Jonsson2020Visual}, and Hamid et al.~\cite{Hamid2019Visual}.

The use of parallel coordinates plots~\cite{Inselberg1987Parallel} is rather prominent for the visualization of automatic hyperparameter tuners such as HyperOpt~\cite{Bergstra2015Hyperopt}. Most of the time, less interactive visualizations have been developed for monitoring automatic frameworks~\cite{Akiba2019Optuna,Golovin2017Google,Jingwoong2018CHOPT,Liaw2018Tune,Liu2019Auptimizer,Tsirigotis2018Orion}. Visualizations arranged into dashboard-styled interfaces are the preferred norm for managing ML experiments and their associated models~\cite{Sung2017NSML,Tsay2018Runway,Wang2020AutoAI,Weidele2020AutoAIViz}.
Automated approaches exclude the user from exploring and refining the hyperparameter search, and the visual representation of automation happens in the form of visualization of the already computed results.


\paragraph*{Human-in-the-Loop Ensemble Learning.} \label{sec:ensem}
There are relevant works that involve the human in interpreting, debugging, refining, and comparing ensembles of models~\cite{Das2019BEAMES,Liu2018Visual,Neto2020Explainable,Schneider2018Integrating,Xu2019EnsembleLens,Zhao2019iForest}. These papers use bagging~\cite{Breiman2001Random} and boosting~\cite{Chen2016XGBoost,Freund1999A,Ke2017LightGBM} techniques for ranking and identifying the best combination of models in different application scenarios. StackGenVis~\cite{Chatzimparmpas2021StackGenVis} is a VA system for composing powerful and diverse stacking ensembles~\cite{Wolpert1992Stacked} from a pool of pre-trained models. On the one hand, we also enable the user to assess the various models and build his/her own ensemble of models. On the other hand, we support the process of generating new models through genetic algorithms and highlight the necessity for the best and most diverse models in the simplest possible voting ensemble. Finally, our approach is model-agnostic and generalizable, since we use both bagging and boosting techniques along with both NNs and simpler models~\cite{Liu2018Visual,Neto2020Explainable,Zhao2019iForest}.

VA tools have also been developed to visualize buckets of models~\cite{Chen2019LDA,Talbot2009EnsembleMatrix,Zhang2019Manifold}, where the best model for a specific problem is automatically chosen from a set of available options. These works feature exploration of the space in search for a final model, but the best model might not be the optimal when compared to a set of models (i.e., multiple hyperparameters) from several algorithms. Additionally, the models are already generated before the exploration, and there is no involvement of an optimization method. 

%% file: 3.Design_goals.tex
In this section, we define the main analytical requirements that a VA system should tackle for supporting evolutionary optimization of hyperparameters.
Then, we describe the corresponding design goals that directed the development of our proposed VisEvol tool. 

\subsection{Analytical Requirements for Evolutionary Optimization}

The analytical requirements (\textbf{R1--R5}) originate from the analysis of the related work in~\autoref{sec:relwo}, including the three analytical needs from Park et al.~\cite{Park2021HyperTendril}, the three key decisions from Wang et al.~\cite{Wang2019ATMSeer}, and the five sub-steps from Li et al.~\cite{Li2018HyperTuner}. 
Also, our own experiences played a vital role, for instance, VA tools for ML such as t-viSNE~\cite{Chatzimparmpas2020t} and StackGenVis~\cite{Chatzimparmpas2021StackGenVis}, and recently-conducted literature reviews~\cite{Chatzimparmpas2020A,Chatzimparmpas2020The}.

\textbf{R1: Identify effective hyperparameters.} 
Interviews performed by Park et al.~\cite{Park2021HyperTendril} showed that users usually sort the models based on a validation metric and then check the hyperparameters of the most performant models (commonly less than 10) for the generated outcomes. 
Next, they select the hyperparameter spaces close to the already explored ones to find more effective hyperparameters using more computational resources. 
However, they cannot be sure whether the updated solution spaces would produce better models, as they might have missed searching for a critical space with better hyperparameters.
Another crucial step is to drop underperforming hyperparameter settings from the candidates or, even better, to reprocess them to become more robust. With crossover and mutation, the final task can be effectively accomplished (see \textbf{R3}).

\textbf{R2: Build an initial ensemble of performant and diverse models.} Wang et al.~\cite{Wang2019ATMSeer} stated that automatic ML approaches yield the model with the highest performance score by default.
Nevertheless, the users could prefer more stable algorithms regarding the adaptations of hyperparameters.
There might even be a case of an underperforming model that could perform better when coupled with another model. 
Those patterns could be lost if only the single best model is available for preview by the user.

\textbf{R3: Send the remaining models for improvement and handle crossover and mutation procedures.} Configuring hyperparameter optimization methods was found unpredictable and disturbing by the interviewees of the investigation by Park et al.~\cite{Park2021HyperTendril}. 
Participants from the interview by Wang et al.~\cite{Wang2019ATMSeer} stated that they revised the hyperparameter space based on previous knowledge.
For this requirement and \textbf{R2}, the users should be able to analyze different algorithms thoroughly and decide to what degree they are going to crossover and mutate each algorithm according to prior knowledge and feedback from VA tools.

\textbf{R4: Contrast the results of all model-generation stages and update the majority-voting ensemble.} In evolutionary optimization, a crossover and mutation phase leads to a propagation of more crossover and mutation phases with exponential growth (cf. \autoref{fig:teaser}(b)). 
Li et al.~\cite{Li2018HyperTuner} found that once the ML expert has acquired all the results from an execution stage, he/she should analyze them with various perspectives and decide if the previously explored models' performance match his/her needs. If not, then more stages should be involved in the process until his/her expectations are met. This entire process should be trackable and manageable from the user's side. The best models (according to the user) are accumulated in a final bucket, forming a majority-voting ensemble.

\textbf{R5: Validate and select a final configuration (single model or combination of models).} 
Automatic ML commonly delivers to users a model with the highest performance according to a single metric (e.g., accuracy), but fails to take into account other characteristics of models~\cite{Park2021HyperTendril}. 
In practice, users want to consider several model features and validation metrics for selecting a model (or models).
Thus, the users want to examine collections of validation metrics and explore model ensembles in various granularities.

\subsection{Design Goals for VisEvol}

To fulfill the described analytical requirements (\textbf{R1--R5}), specifically in the context of VA and ensemble learning, we have derived five design goals (\textbf{G1--G5}) to be tackled by our tool. The implementation of these design goals is described in \autoref{sec:overview}.

\textbf{G1: Analysis of predictions and validation metrics for the identification of effective hyperparameters.} 
We aim to support the exploration of algorithms and models with various hyperparame\-ters (\textbf{R1}) as follows:
(1) illustrate the performance of each algorithm and model based on multiple validation metrics chosen by the user; 
(2) project the models into a hyperparameter embedding according to the previous overall performance using DR methods; (3) compare the mean performance of all algorithms and models vs. a selection of models for every metric; and (4) analyze the predictive results for each instance and for all models against a selection of models with regard to the difference in predictive power.

\textbf{G2: Migration of powerful and alternative models to the majority-voting ensemble.} In continuation of the preceding goal, our VA tool should allow the users to pick the best (and most diverse) models for the ensemble from different areas in the projection (\textbf{R2}). 
Using the other coordinated views, the user can compare the selected models against all models and act accordingly.

\textbf{G3: Transformation of underperforming models via crossover and mutation.} Users are able to conduct similarity-based analyses relevant to the algorithms' predictions, which are the initial indications for tuning crossover and mutation of Stage~1 ($S_1$). The various perspectives that multiple views have to offer for several algorithms further support those prior analyses. 
We aim to provide explicit visual feedback from $S_1$ of crossover and mutation toward $S_2$ for users to select appropriate numbers for model optimization and test their hypotheses (\textbf{R3}).

\textbf{G4: Comparison of multi-stage generated hyperparameter sets in various granularities.} An addition to \textbf{G1} is that the positive or negative impact of performance should be measured during the creation of models through the multi-stage crossover and mutation procedure.
\hl{VisEvol should thus display both successful and underperforming paths for every crossover and mutation stage (\textbf{R4}).}

\textbf{G5: Extraction of an ultimate model or a voting ensemble with a side-by-side performance comparison.} A comparison between the currently active ensemble against the optimal solution found until that point in time should be established in our tool to assist the extraction of a competitive and effective ensemble (\textbf{R5}). 

%% file: 4.System_overview.tex
%
Following our analytical requirements and derived design goals, we have developed VisEvol, an interactive web-based VA tool that allows users to utilize evolutionary optimization in order to search for effective hyperparameters.
It is implemented in JavaScript using the Vue.js~\cite{vuejs} framework and a combination of D3.js~\cite{D3} and Plotly.js~\cite{plotly} visualization libraries for the frontend.
The backend is implemented in Python using Flask~\cite{Flask} as the web framework and Scikit-Learn~\cite{Pedregosa2011Scikit} as the ML library. 

\begin{figure}[t]
\centering
\includegraphics[width=\linewidth]{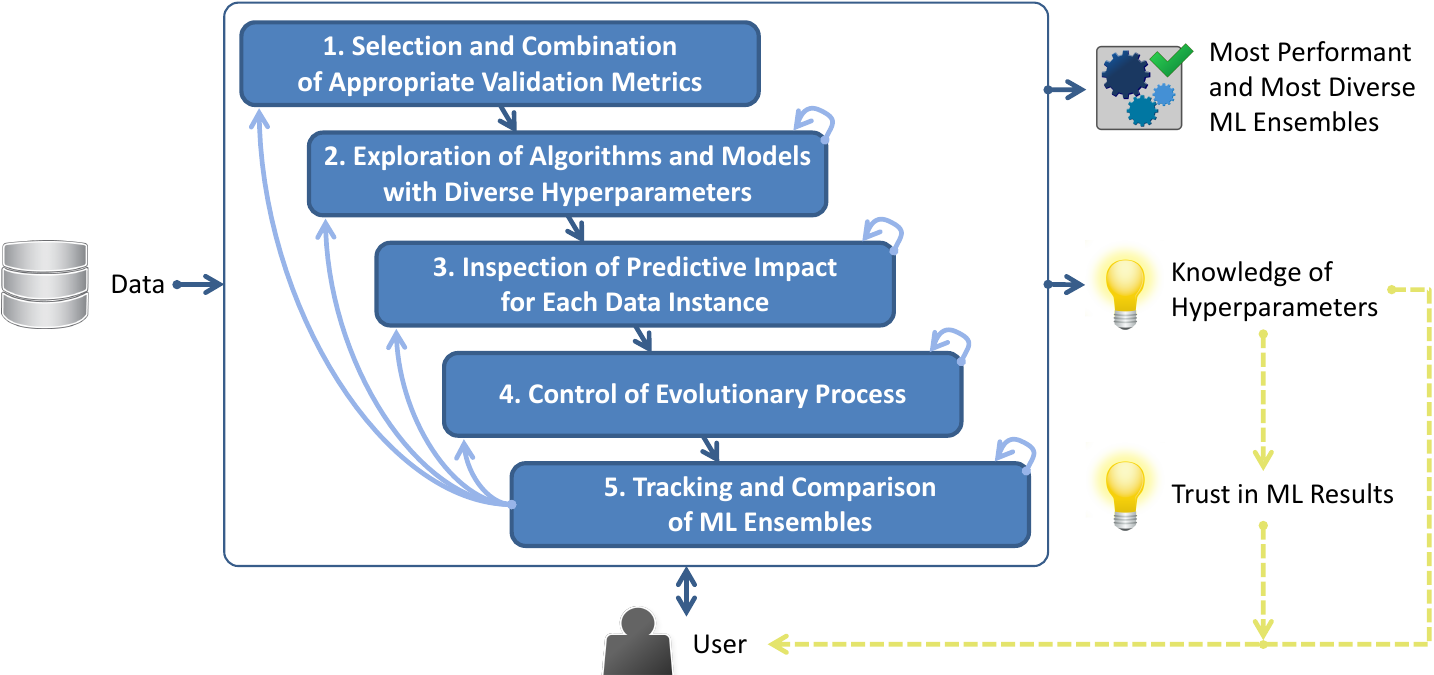}
\caption{The VisEvol workflow allows the users to construct performant and diverse ML ensembles, gain knowledge about the hyperparameters chosen via the evolutionary optimization process, and thus gain trust in the respective ML results. \hl{The users are capable of interacting with all phases iteratively, represented by the multiple arrows inside the box.}}
\label{fig:workflow-diagram}
\end{figure}

The tool consists of eight main interactive visualization panels (\autoref{fig:teaser}): (a) data sets and validation metrics ($\rightarrow$  \textbf{G1}), (b) process tracker and algorithms/models selector ($\rightarrow$  \textbf{G3}), (c) overall performance for each algorithm/model, (d) hyperparameter space, (e) majority-voting ensemble ($\rightarrow$  \textbf{G2}), (f) performance for each validation metric, (g) predictive results for each data instance ($\rightarrow$  \textbf{G4}), and (h) performance for majority-voting ensemble ($\rightarrow$  \textbf{G5}). 
We propose the following \textbf{workflow} for the integrated use of these panels (cf. \autoref{fig:workflow-diagram}): (i) choose suitable validation metrics for the data set, which are then used for validation during the entire process (\autoref{fig:teaser}(a));
(ii) in the next exploration phase, compare and choose specific ML algorithms for the ensemble and then proceed with their particular instantiations, i.e., the models (see \autoref{fig:teaser}(c--e));
(iii) during the detailed examination phase, zoom in into interesting clusters already explored in the previous phase, and focus on indications that confirm either their approval in the ensemble or their need for transformation through the evolutionary process (cf. \autoref{fig:teaser}(f and g)); 
(iv) control the evolutionary process by setting the number of models that will be used for crossover and mutation in each algorithm (\autoref{fig:teaser}(b)); and 
(v) compare the performances of the \emph{best} so far identified ensemble against the \emph{active} majority-voting ensemble in \autoref{fig:teaser}(h). This is an iterative process with a final composition of the most performant and most diverse majority-voting ensemble. The generated knowledge regarding hyperpara\-me\-ters is fed back to the user, whose trust in the results increases, and he/she stops when his or her expectations are met. The individual panels and the workflow are discussed in more detail below.
%

The user interface of VisEvol is structured as follows: 
(1) two projection-based views, referred to as \emph{Projections 1 and 2}, occupy the central UI area (cf. \autoref{fig:teaser}(d and e));
(2) active views relevant for \emph{both} projections are positioned on the top (cf. \autoref{fig:teaser}(b and c)); and
(3) commonly-shared views that update on the exploration of \emph{either} Projection 1 \emph{or} 2 are placed at the bottom (see \autoref{fig:teaser}(f and g)).
Thus, \autoref{fig:teaser}(h) is always active for Projection 2, as it is related to the majority-voting ensemble.
\emph{Soft majority voting} strategy (i.e., predicted probabilities) is always applied. 

\hl{To exploit the model-agnostic nature of our proposed workflow, VisEvol supports five different supervised ML algorithms (any could have been used):  
(1) a neighbor classifier (\emph{k-nearest neighbor (KNN)}), (2) a linear classifier (\emph{logistic regression (LR)}), (3) an NN classifier (\emph{multilayer perceptron (MLP)}), and (4) two ensemble classifiers (\emph{random forest (RF)} and \emph{gradient boosting (GradB)}).}
The primary hyperparameters used for mutation: \emph{number of neighbors} for KNN, \emph{inverse of regularization strength} for LR, \emph{hidden layer sizes} for MLP, \emph{number of decision trees} for RF, and \emph{number of boosting stages} for GradB.

In the following subsections, we explain the system by using a running example with the \emph{heart disease} healthcare data set obtained from the UCI Machine Learning repository~\cite{Dua2017}. 
The data set represents a binary classification problem and consists of 13 numerical features/attributes and 303 instances. It is rather balanced, with 138 patients being healthy and 165 having a diseased heart.

\begin{figure*}[tb]
\centering
\includegraphics[width=\linewidth]{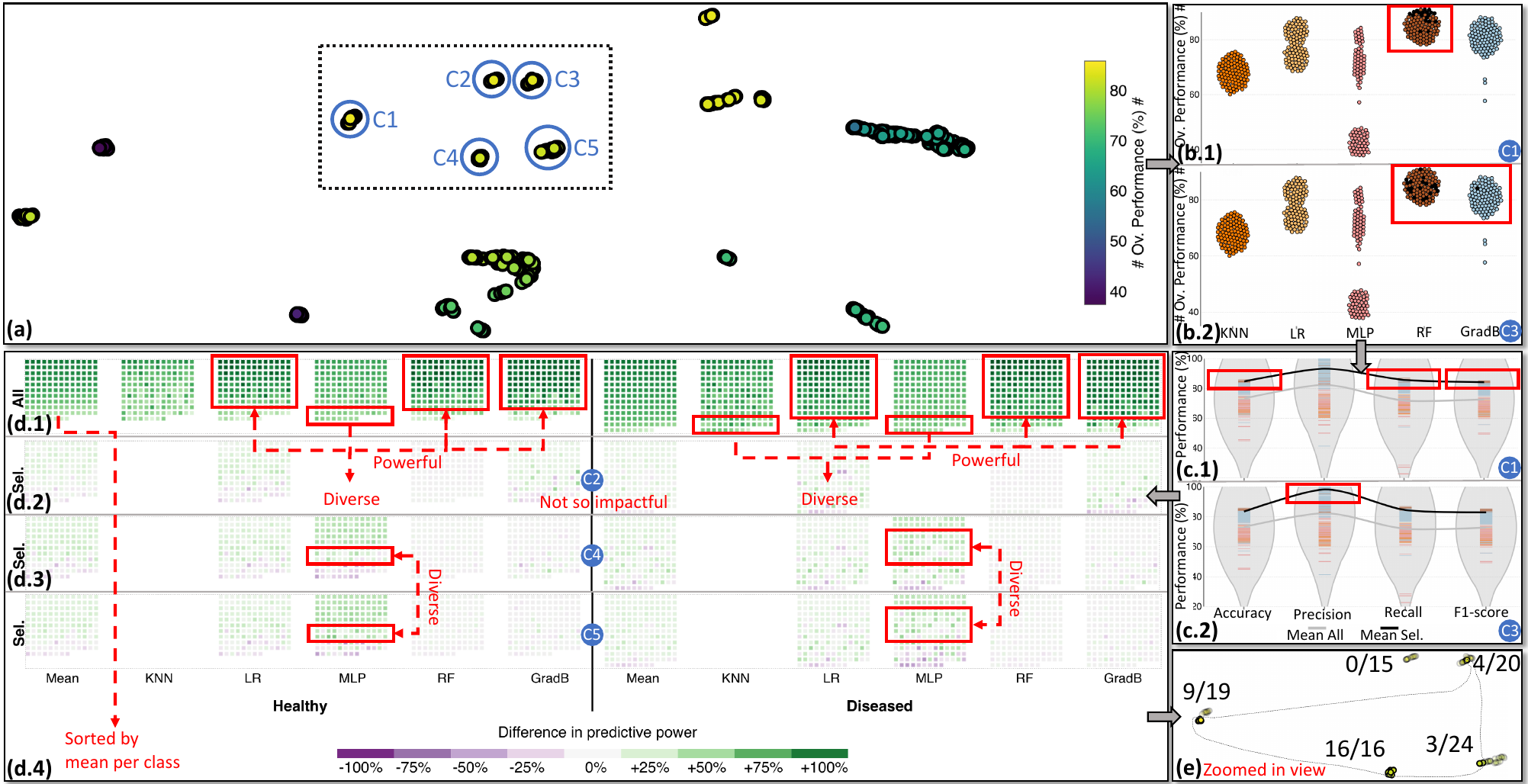}
\caption{Exploration of ML models with VisEvol.
View (a) presents a selection of similar and better-performing models in several clusters.
(b.1) indicates that \circled{C1} contains well-performing RF models, in contrast to (b.2), in which \circled{C3} includes more diverse RF models and a GradB model. 
For the accuracy, recall, and f1-score metrics, \circled{C1} performs much better than the average, based on the bean plots in (c.1). 
However, \circled{C3} achieves better results for the precision metric.
In the grid-based view (d.1), LR, RF, and GradB algorithms appear more powerful than other algorithms that are more diverse due to the good predictions of hard-to-classify instances. 
\circled{C2} seems redundant because of the \circled{C4} and \circled{C5} that improve similar cases (d.2). If we look at (d.3) and (d.4), both visualizations display MLP models that enhance the predictive power of different instances in both classes. \hl{Finally in view (e), we mix models from the multiple explored clusters to create the first voting ensemble.}}
\label{fig:use_case1_model_s0}
\end{figure*}

\subsection{Data Sets and Validation Metrics} \label{sec:metrics}

\hl{Support for (1) selecting proper validation metrics for balanced and imbalanced data sets and (2) directing the experts' attention to different classes for the given problem constitute two of the critical open challenges in ML.} 
For instance, accuracy is preferred to the g-mean metric for a balanced data set~\cite{Bekkar2013Evaluation}. 
\hl{In another example, a medical expert might focus more on eliminating false-negative predictions than false-positives (e.g., a patient being actually ill but predicted as healthy) with a bad impact on the latter. However, this trade-off is necessary when considering a person's life.} 

In VisEvol, up to eight different metrics can be used simultaneously, depending on the number of instances falling into each class of a binary classification problem. 
The available metrics are divided into two groups: balanced data sets ($\rightarrow$ \emph{accuracy}, \emph{precision}, \emph{recall}, and \emph{f1-score}) and imbalanced data sets ($\rightarrow$ \emph{g-mean}, \emph{ROC AUC}, \emph{log loss}, and \emph{MCC}). 
For the initialization of VisEvol, the user should direct his/her attention to the top-left panel shown in \autoref{fig:teaser}(a); the preferrable group of validation metrics will depend on the distribution of instances in the two classes for each individual data set (Step 1 in \autoref{fig:workflow-diagram}). 
\hl{Then he/she sets a number of models $n$ with the slider shown in \autoref{fig:teaser}(a), from 50 to 300, which will impact the initial search of random hyperparameters.}
Choosing a value $n$ is a matter of finding a balance between spending computational resources and time against scanning more accurately the solution space for better hyperparameter tuples.
The $k$ value is used for the k-fold cross-validation, with the options of 5, 10, or 15 folds. 

In our running example, knowing that the data set is balanced, we use the first group of metrics (currently unselected in \autoref{fig:teaser}(a)), and leave the default random search to 100 models per algorithm, leading to 500 models in total. The $k$ value is set to 10 because we want to precisely compare our results with recent work~\cite{Latha2019Improving}. 

\subsection{Hyperparameter Space} \label{sec:algs}

\hl{To provide a holistic view on the performance of the models for the selected validation metrics, we use a UMAP~\cite{McInnes2018UMAP} projection, as seen in \autoref{fig:use_case1_model_s0}(a), that consists of the 500 randomly-sampled models (MDS~\cite{Kruskal1964Multidimensional} and t-SNE~\cite{vanDerMaaten2008Visualizing} are also available).}
%
\hl{Each model uses a set of particular hyperparameters, and it is projected from the space of validation metric values (here 4 dimensions, but could be more).} 
Thus, groups of points represent clusters of models that perform similarly according to all the metrics. 
The plot uses the Viridis colormap~\cite{Liu2018Somewhere} to show the average performance of each model according to all selected metrics. 
This view provides the user with an overview of the hyperparameter space and ability to look for previously-unknown patterns. 
We can now select high-performing clusters and proceed with deciding which models to include in our ensemble (Step 2 in \autoref{fig:workflow-diagram}).

\hl{At this phase, we want to confirm precisely the cluster affiliation and the relationship with the overall performance (here, the average of 4 validation metrics) for all the models.} To achieve that, the beeswarm plots in~\autoref{fig:use_case1_model_s0}(b.1 and b.2) arrange the models according to the distinct algorithms in the x-axis and sort the models based on the overall performance along the y-axis (abbreviated to \emph{Ov. Performance}).
Similarly-performing models can overlap in this view (due to the y-axis values), so we apply a force-based layout algorithm to make sure they are visible.
However, moving models around introduce uncertainty in this view. Thus, we visualize the mean deviation in pixels for every algorithm (cf. \autoref{fig:teaser}(c), bar chart at the bottom) to minimize any misleading visualization bias. \autoref{fig:use_case1_model_s0}(b.1) suggests that \circled{C1} contains the most performant RF models, while \autoref{fig:use_case1_model_s0}(b.2) presents less performant RF and GradB models situated in \circled{C3} that may, however, be more diverse (black dots represent selected models). 

\begin{figure*}[t]
  \centering
  \includegraphics[width=\linewidth]{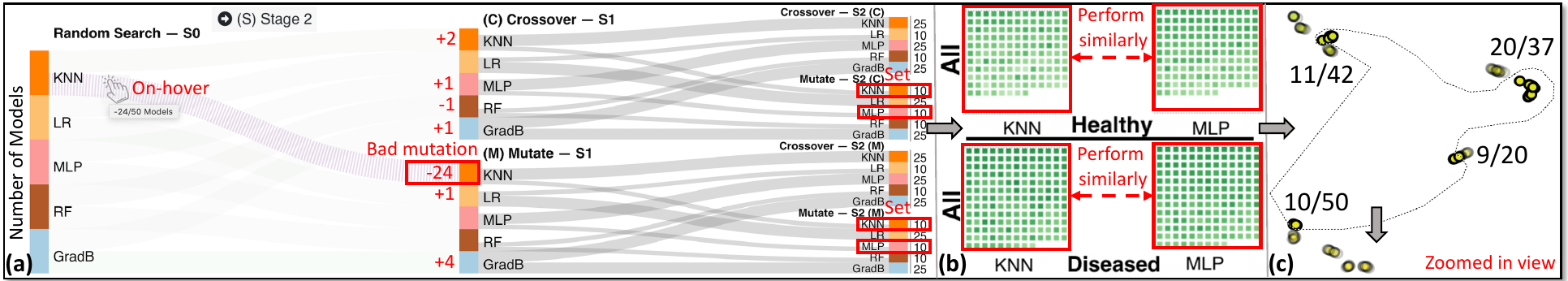}
  \caption{Tuning the crossover and mutation process toward $S_2$. In (a), we set fewer models for mutation and more for crossover for both KNN and MLP algorithms. Our choice is motivated by the feedback received from the bad KNN mutation in $S_1$ and the fact that KNN and MLP perform almost identically for both independent classes (as illustrated in (b)). Similar to \autoref{fig:use_case1_model_s0}, we investigate clusters in the projection, select a few models from each explored cluster shown in (c), and send the rest for crossover and mutation.}
  \label{fig:use_case1_model_s1}
\end{figure*}

\begin{figure*}[tb]
  \centering
  \includegraphics[width=\linewidth]{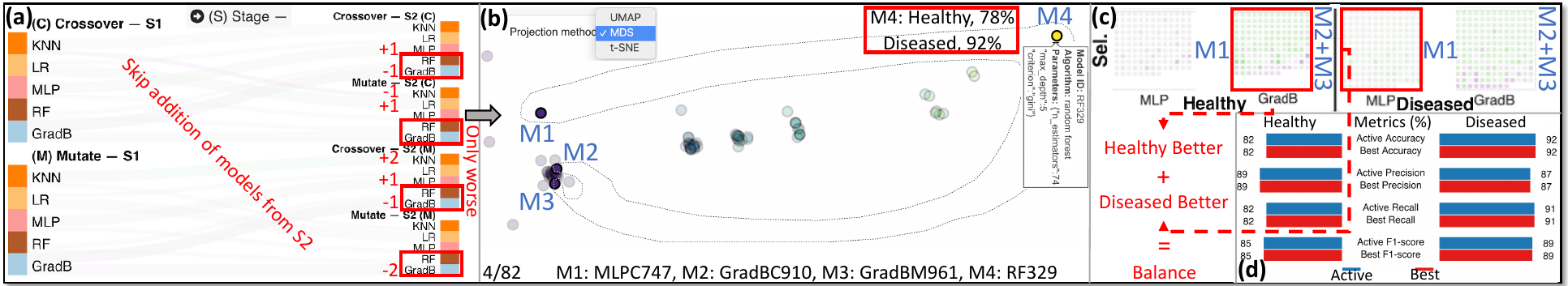}
  \caption{The outcome of the $S_2$ evolutionary optimization procedure and the final voting ensemble: \hl{(a) highlights that we have reached an impactful solution, since the models are not getting significantly better.} Thus, we skip the addition of models from $S_2$. In (b), after an extensive exploration of the majority-voting ensemble, we end up with the selection of four models: M4 originating from the initial random search (the most performant when used individually) and M1--M3 from the crossover and mutation processes at $S_1$. We narrow down this selection even further by examining (c), where one MLP model appears to perform better for the Diseased class and two GradB models for the Healthy class. 
  The active performance matches the best performance found so far (d). Hence, this is the most powerful majority-voting ensemble.
  }
  \label{fig:use_case1_model_s2}
  
\end{figure*}

\hl{For clusters \circled{C1} and \circled{C3}, we want to find out the relation with the different validation metrics.} 
\hl{This task is supported by the bean plots in ~\autoref{fig:use_case1_model_s0}(c.1 and c.2), which are separated on the x-axis by the selected metrics from the validation metrics panel (see \autoref{fig:teaser}(a)).} 
The \emph{beans} (lines inside each bean plot) represent all the individual models. However, they were not designed with a goal to accurately represent each individual model, since there are cluttering problems due to the thickness of the line for each bean and the large number of models per bean plot. The main purpose of this plot is to check if the mean values for a selection are better when compared to the overall mean, as shown in~\autoref{fig:use_case1_model_s0}(c.1). For \circled{C3} (\autoref{fig:use_case1_model_s0}(c.2)), the precision metric is further improved compared to \circled{C1}.

At this point, the importance of \circled{C1} and \circled{C3} is clear, so we decide to gradually scan for the in-depth connections of the models belonging to the remaining clusters and the data instances (Step 3 in \autoref{fig:workflow-diagram}). \hl{The grid-based visualization in~\autoref{fig:use_case1_model_s0}(d.1) focuses on the exploration of the predictive power (0\%--100\%) over the data set's instances represented with white to dark green colors.}
If a data set contains fewer than 169 instances per class, then we display all of them in the grid. 
Otherwise, in order to scale the visualization to larger data sets (e.g., our next use case), we first build a grid with a fixed number of cells (100). Then, we use \emph{K-means clustering} to place the data set's instances within these cells, i.e., we create 100 clusters, one per cell of the grid. 
This action is noticeable due to specific play/stop glyphs, as illustrated in~\autoref{fig:teaser}(g), top-left corner.
There is one grid per different ML algorithm, plus one grid for the overall values (leftmost grid in \autoref{fig:use_case1_model_s0}(d.1)), and all the instances are sorted according to the mean performance for all the explored models.
\hl{Each cell of the grid (as shown in \autoref{fig:use_case1_model_s0}(d.2--d.4)) then presents the computed difference in predictive power for all its instances (from $-$100\% to $+$100\%) for the selected against all models.} The color-encoding diverges from purple to green for negative to positive difference.
In case the K-means clustering functionality is active, we use bar charts to depict the distribution of instances in the 100 individual cells (see \autoref{fig:teaser}(g), bottom). Afterwards, the predictive power for every cell is computed on average from all the instances that belong to it.

\hl{From \autoref{fig:use_case1_model_s0}(d.1), we observe that KNN and MLP contain more diverse models (darker green color for instances at the bottom), because they better predict hard-to-classify instances when compared to LR, RF, and GradB (which work better for the easy-to-predict instances).} Since we have already found powerful and diverse RF and GradB models from the prior analyses in \circled{C1} and \circled{C3}, we now focus the MLP algorithm. For \circled{C2}, we were unable to find any impactful models, i.e., only a tiny amount of instances are green-colored in \autoref{fig:use_case1_model_s0}(d.2). Models originating from \circled{C4} and \circled{C5} sufficiently cover the need for diversity, including MLP models. 
We then pick the models shown in~\autoref{fig:use_case1_model_s0}(e) for our voting ensemble, and the remaining models are updated by the evolutionary optimization (see next subsection). This action concludes $S_1$. 

\subsection{Process Tracker and Algorithms/Models Selector}

After the initial generation of hyperparameter settings with the use of random search ($S_0$), we manually evaluated the models and forwarded the remaining unselected models for crossover and mutation~\cite{Cantu2005An}. As explained in~\autoref{sec:intro}, crossover blends randomly different models from the same algorithms (or else it is impossible due to differentation in hyperparameters), and mutation captures the primary hyperparameter (previously mentioned in~\autoref{sec:overview}) and randomly mutates it with new values, which were previously not explored. This procedure repeats for every algorithm separately.

In the Sankey diagram (see \autoref{fig:use_case1_model_s1}(a)), the user tracks the progress of the evolutionary process and is able to limit the number of models that will be generated through crossover and mutation for each algorithm (Step 4 in \autoref{fig:workflow-diagram}). The default here is defined as \emph{user-selected random search value / $2$} for each algorithm, to sustain the vertical symmetry in the Sankey diagram, as shown in \autoref{fig:use_case1_model_s1}(a), left. For $S_1$, we choose to keep the default values for crossover and mutation, but an analyst with prior knowledge and experiences could fine-tune this process. While moving toward $S_2$, we notice from~\autoref{fig:use_case1_model_s1}(b) that KNN and MLP perform similarly. The output of $S_1$ in~\autoref{fig:use_case1_model_s1}(a) becomes the input for $S_2$, assisting us in the selection of appropriate numbers for model generation for $S_2$. When we hover over a path of the Sankey diagram, we see how many models perform better or worse than the already-explored models for each particular algorithm. The color-encoding is the same as~\autoref{fig:use_case1_model_s0}(d.2--d.4)), and it is measured as the \emph{number of overperforming models compared to the initial models / total crossover or mutate models} for each algorithm. If there are no overperforming models, then we show \emph{number of underperforming models compared to the initial models / total crossover or mutate models} for each algorithm. This approach primarily allows the user to identify how many models are improved based on each transformation (crossover or mutation), but it also highlights cases with very bad results from crossover or mutation, where no better ML models could be found. In our example, KNN mutation produced bad results, hence, we set the subsequent KNN and MLP mutations (due to the previously-discussed similarity in~\autoref{fig:use_case1_model_s1}(a)) to lower values than the default ($10$ vs. the default of $25$). The visualization reduces the width of each path line in the Sankey diagram accordingly when the values are smaller than the maximum permitted. Next, we apply an equivalent procedure for all the algorithms. Finally, an analysis is conducted in a similar way to previous sections, with the selection of points in~\autoref{fig:use_case1_model_s1}(c) as an outcome.


\subsection{Majority-Voting Ensemble}
\label{subsec:majority-voting-ensemble}

From~\autoref{fig:use_case1_model_s2}(a), right, we see that only a few KNN, LR, and MLP models were better than the previous stages. \hl{Thus, we conclude that there is no further improvement, and it is hard to find better hyperparameter tuples.} We skip the addition of models from $S_2$ to the final ensemble because RF and GradB seem to perform better overall. In \autoref{fig:use_case1_model_s2}(b), we switch the embedding to an MDS projection which favors the global structure, and compare clusters of models until we discover the active ensemble that contains M1--M4 models (using the \emph{Compute performance for active ensemble} button present in~\autoref{fig:teaser}(e)). \autoref{fig:use_case1_model_s2}(c) suggests that M2+M3 are better for the \emph{Healthy} class, while M1 is better for the \emph{Diseased} class. M4 is somewhere in-between but very powerful overall. By keeping the balance in this ensemble, we achieve the highest recorded performance for our analysis (cf. horizontal bar chart in~\autoref{fig:use_case1_model_s2}(d)). The symmetric horizontal bar chart is split vertically based on the different metrics. The left-side is about one target class and the right side for the other one. Blue is always used for the actively explored ensemble combination while red is for the best ensemble found yet. The comparison of both serve the purpose of identifying exceptionally performing majority-voting ensembles (Step 5 in \autoref{fig:workflow-diagram}).

\subsection{Performance of Majority-Voting Ensemble}

Latha and Jeeva~\cite{Latha2019Improving} tried out various ensembles for this same data set, with or without (as in our case) feature selection. They found that applying majority vote with the NB, BN, RF, and MLP algorithms was the best combination, achieving $\approx$82\% accuracy without feature selection. However, they do not state how many models were used in the composition of this ensemble.
With Vis\-Evol, we reached an accuracy of 87\% with only 4 ML models (see \autoref{fig:use_case1_model_s2}(d)), thus surpassing their majority-voting ensemble. If the user wants to utilize one model, our selection would have been M4:RF329 (see \autoref{fig:use_case1_model_s2}(b), top-right), which has a combined predictive accuracy of 85\%.
This shows that our VA approach can be effective in searching for hyperparameters and building powerful, simple, and diverse voting ensembles.  

%% file: 5.Use_case.tex
\begin{figure*}[t]
  \centering
  \includegraphics[width=\linewidth]{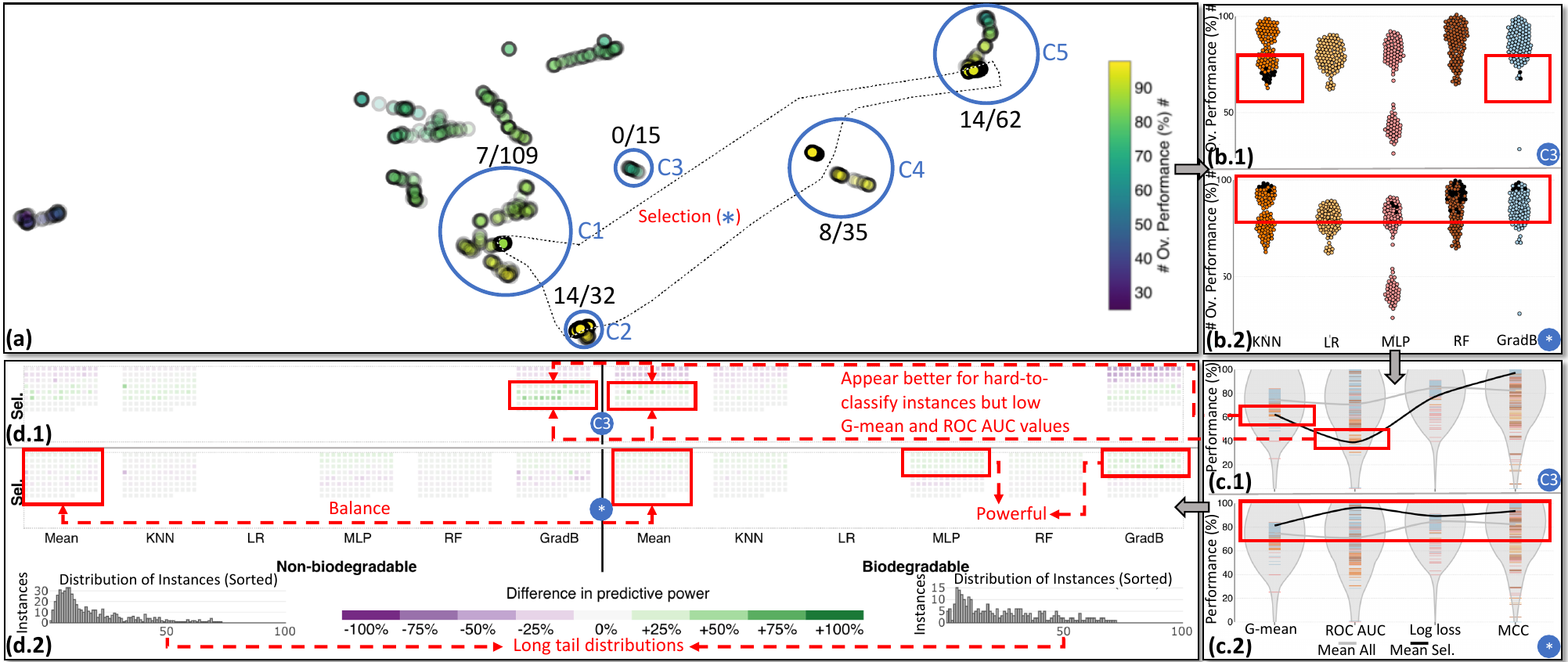}
  \caption{The exploration of clusters of interest that contain performant ML models. View (a) presents the user's selection that drive the analyses performed in the remaining subfigures. (b.1) provides an overview of the performance, showing that \circled{C3} has underperforming KNN and GradB models. On the other hand, (b.2) shows that the user's choice of models retains both performance and diversity. In (c.1), we observe that g-mean and ROC AUC scores are very low, which is a problem investigated further in view (d.1). Those models appear to perform better for the hard-to-classify instances; however, this is a misconception. (c.2) gives supporting evidence to the user's selection, since all validation metrics are higher than the average values for all models, along with the in-depth visualization in (d.2).}
  \label{fig:use_case2_S1}
  \vspace{-1em}
\end{figure*}

\begin{figure*}[tb]
  \centering
  \includegraphics[width=\linewidth]{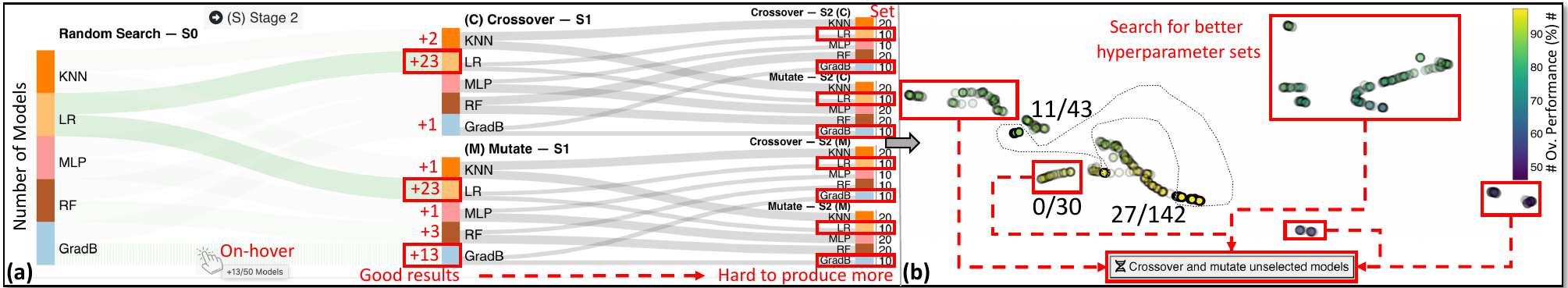}
  \caption{Setting crossover and mutation values for $S_2$ evolutionary optimization. The Sankey diagram's feedback in (a) suggests that new hyperparameter sets that perform better than the current models will be hard to produce. Despite that, after user's selection in (b), the leftover models crossover between each other and mutate over successive evolutions.}
  \label{fig:use_case2_S2}
  \vspace{-1em}
\end{figure*}

In this section, we describe how VisEvol can be used to improve the results of a study about the relationships between \emph{chemical structure and biodegradation of molecules}, when compared to previous work from Mansouri et al.~\cite{Mansouri2013Quantitative}. 
The \emph{QSAR (Quantitative Structure Activity Relationships) biodegradation} data set represents a binary classification problem where molecules are assigned to either the \emph{Biodegradable} or \emph{Non-Biodegradable} classes. \hl{The class distribution is rather imbalanced, with 284 degradable and 553 non-degradable molecules for the training set that contains 41 diverse features.}
For their solution, the authors trained three ML models (KNN, PLSDA, and SVM) and then combined their results using two consensuses.
%
We contrast their first consensus with our majority-voting ensemble, and we compare our results with the same validation metrics for the unseen test and external validation data, which simulate a real-world situation. 

\textbf{Exploration and Selection of Algorithms and Models.} Similar to the workflow described in \autoref{sec:overview}, we start by setting the most appropriate validation metrics for the imbalanced data set (see~\autoref{fig:teaser}(a)). The projection in~\autoref{fig:use_case2_S1}(a) offers an overview of the high performing clusters that need further investigation (\circled{C1}--\circled{C5}). By looking at~\autoref{fig:use_case2_S1}(b.1), we infer that \circled{C3} contains KNN and GradB models that perform worse than the remaining models, in general. For \circled{C3}, \emph{MCC} and \emph{log loss} are very high when compared to the \emph{g-mean} and \emph{ROC AUC} metrics, as shown in~\autoref{fig:use_case2_S1}(c.1). This can be further explained if we dig into this cluster's performance for each instance. Those models create an illusion of performing well for the hard-to-classify instances (\autoref{fig:use_case2_S1}(d.1)). Nevertheless, the previously spotted low values in \emph{g-mean} and \emph{ROC AUC} suggest that those models reach high precision but with low recall, or vice versa; hence, they should be avoided. On the contrary,~\autoref{fig:use_case2_S1}(b.2) presents a blend of performant models, reaching an equilibrium state where all the values for the metrics are concurrently high (cf. \autoref{fig:use_case2_S1}(c.2)). The explored GradB models, for example, improve the \emph{Biodegradable} class and accomplish a balance between the two mean values of the bins for both classes (see \autoref{fig:use_case2_S1}(d.2)). This is especially true when we observe that the distributions of the instances (based on 100 clusters generated by KNN) are long-tailed. That means most instances belong in the first sorted bins, which are predicted better than the following bins (as described in \autoref{sec:overview}).

\textbf{Tuning the Evolutionary Optimization Process.} After $S_1$'s default execution with 100 models for each algorithm (50 due to crossover and 50 because of mutation), we continue with setting the next batch of crossover and mutation processes. We received useful feedback from $S_1$ that supports us in creating new models for $S_2$ in \autoref{fig:use_case2_S2}(a), left. When we hover over a mutation path for GradB, we see that 13 out of 50 models perform better than the initial ones from random search. This could mean that it is hard to produce new models that will outperform the previous production of stable and robust models. The same applies to the LR model, for both crossover and mutations paths (with 23 out of 50 models). 
Thus, we choose to set the production from 25 models to 10 for both LR and GradB algorithms. 
Similarly to the previous paragraph, we select well-performing and diverse models as shown in \autoref{fig:use_case2_S2}(b), and the unselected models are being used in the crossover and mutation method based on the previously adjusted parameters.

\textbf{Examining the Influence of Diversity.} \hl{In \autoref{fig:teaser}(b), we see that most paths fail to create extra powerful models, indicating that it will be hard to find better models.} The GradB algorithm seems to have generated a few enhanced performance models from the crossover path to mutation at $S_2$ (light-green color).
After another hyperparameter space search (see \autoref{fig:teaser}(d)) with the help of supporter views (\autoref{fig:teaser}(c, f, and g)), out of the 290 models generated in $S_2$, we select 28 to add to the ensemble (cf. \autoref{fig:teaser}(e)). 
Surprisingly, the best majority-voting ensemble for the test and validation data sets contains 1 RF and 3 GradB models, compared to the 110 models added from all stages in total.
This currently active ensemble appears to perform worse in the 5-fold cross-validation results against the current best ML ensemble (\autoref{fig:teaser}(h)). Though, this could imply that a larger $k$-fold value should have been used from the start.
%
However, for comparison purposes, we have chosen the $k$ value of 5 from Mansouri et al.~\cite{Mansouri2013Quantitative}. 

\textbf{Evaluation with the Test and External Validation Sets.} To verify whether our findings were reliable, we applied the resulting majority-voting ensemble to the same test and external validation data sets as Mansouri et al.~\cite{Mansouri2013Quantitative}, see \autoref{tab:results}. For the test data set, the reported accuracy was approximately 87\%.
In our case, we reached 89\% for accuracy with the final voting ensemble (\emph{macro-average}). 
Additionally, as an extra validation, we checked the results for the additional external data set.
Using our approach, we managed to achieve the same accuracy as before, 89\%, compared to 83\% reported by Mansouri et al.~\cite{Mansouri2013Quantitative}.
%

\begin{table}[h!]
  \centering
  \caption{Summary of the test data and external validation data results for the QSAR biodegradation binary classification problem.}
  \label{tab:results}
  \includegraphics[width=\columnwidth]{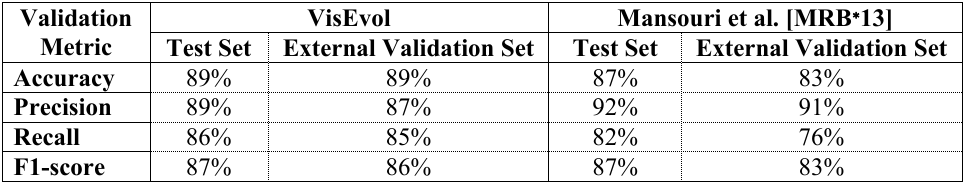}
\end{table}
\vspace{-2mm}

%% file: 6.Evaluation.tex
%
\hl{We conducted three online semi-structured interviews with ML experts to obtain qualitative feedback about our tool's usefulness, as in prior works~\cite{Ma2020Explaining,Xu2019EnsembleLens}.}
The first expert (\textbf{E1}) is a senior lecturer in mathematics working with reinforcement learning and has approximately 3 years of experience with ML. He recently acquired his PhD in mathematics and has basic knowledge regarding ensemble learning. The second expert (\textbf{E2}) is a senior researcher in software engineering and applied ML working in a government research institute and as an adjunct professor. He has worked with ML for the past 7.5 years, and 2.5 years with ensemble learning. The third expert (\textbf{E3}) is an ML engineer and manager in a large multinational company, working with recommendation systems. She has approximately 7.5 years of experience with ML, of which 2 years are associated with ensemble learning. \hl{The latter two experts have PhDs in computer science; none of our three experts reported colorblindness issues.} The followed procedure was: (1) presentation of the key goals of VisEvol, (2) demonstration of the functionality of each view and interaction with the tool using the \emph{heart disease} data set, and (3)  explanation of the process of boosting the results in~\autoref{sec:overview}.
Each interview took about one hour. 
We informed the participants of the main areas we expected feedback from, but they were free to comment on anything.

\textbf{Workflow.}
\textbf{E1} and \textbf{E2} commented that the workflow of VisEvol is well designed. Although \textbf{E3} expected a more linear workflow, she agreed that the combined views are better positioned at the top, with the interactive projections in the middle and the shared views at the bottom. \textbf{E2} has recently worked with genetic algorithms for testing traffic-scenarios for autonomous vehicles. In that case, they had to set a strict budget before execution and perform multiple crossover and mutation stages which can take days to run. \hl{Nevertheless, he noticed that in evolutionary optimization, hundreds of stages might not be necessary since, with three stages, we could gather performant models that are hard to surpass in terms of predictive performance.} Finally, \textbf{E1} mentioned that controlling the evolutionary process via the Sankey diagram can be time-saving.

\textbf{Visualization and Interaction.} 
\textbf{E1} and \textbf{E3} were delighted by the possibilities of the visual exploration of the hyperparameter space. \textbf{E1} was enthusiastic about the grid-based view and stated that it is a game-changer for finding performant and diverse models. 
\textbf{E2} was initially confused with the comparison of instances from various clusters in this same view, but after some training period, he understood that he has to look for different patterns collectively (instead of individual instances). Afterwards, he agreed that, once he gradually explored a cluster, it was easier to gain insights from the comparison with the rest. Both \textbf{E1} and \textbf{E2} mentioned that even though interactions are mostly bounded in the projection-based views, this keeps the tool easy to interact with and removes additional complexity, which is an excellent practice to follow with people not accustomed to VA tools. 
``It is great to find that various combinations of models lead to different ensembles that are better for each class of the independent variable, which is visible from the two views [in~\autoref{fig:teaser}(g) and (h)]'', said \textbf{E2}. Although this extra information restricts the generalization of VisEvol in non-binary classification problems, modifying those views should be straightforward. 

\textbf{Limitations.}
\textbf{E1} and \textbf{E2} were worried about the \emph{scalability} of the tool. Indeed, the excessive computational time required for producing new hyperparameters along with ensemble learning methods can be problematic. \hl{Despite that, one possible improvement for VisEvol is to utilize parallel processing on powerful cloud servers. Moreover, we believe that the advancements in high-performance hardware and progressive VA and data science workflows~\cite{Stolper2014Progressive,Turkay2018Progressive} will be beneficial for VisEvol. 
The users can also avoid extra computations at certain steps of majority-voting ensemble construction, as discussed in Section~\ref{subsec:majority-voting-ensemble}}. 
Another open issue is the \emph{avoidance of hyperparameter tuning per se}, as noted by \textbf{E3}. \hl{The goal of the tool is not to explore or bring insights about the individual sets of hyperparameters of the models or algorithms, but instead we focus on the search for new powerful models and implicitly store their hyperparameters.}
The study of the impact of particular hyperparameters is considered as a future direction for VisEvol. Also, \textbf{E3} stated that we could allow the user to \emph{specify the hyperparameters range} at every stage and test \emph{alternative mutation strategies}~\cite{Cantu2005An}. \textbf{E1} expressed his interest in \emph{checking combinations of evolutionary optimization} with the crossover and mutation process applied to the best-performing models (e.g.,~\cite{Young2015Optimizing}). However, as the user usually adds---as few as possible---models to the ensembles, the hyperparameters' evolution for the excluded algorithms will be infeasible. We plan to overcome such limitations. 

%% file: 7.Conclusion.tex
In this paper, we presented VisEvol, a VA tool with the aim to support hyperparameter search through evolutionary optimization. With the utilization of multiple coordinated views, we allow users to generate new hyperparameter sets and store the already robust hyperparameters in a majority-voting ensemble. Exploring the impact of the addition and removal of algorithms and models in a majority-voting ensemble from different perspectives and tracking the crossover and mutation process enables users to be sure how to proceed with the selection of hyperparameters for a single model or complex ensembles that require a combination of the most performant and diverse models. The effectiveness of VisEvol was examined with use cases using real-world data that demonstrated the advancement of the methods behind achieving performance improvement. Our tool's workflow and visual metaphors received positive feedback from three ML experts, who even identified limitations of VisEvol. These limitations pose future research directions for us.

%% file: VisEvol.bbl
\newcommand{\etalchar}[1]{$^{#1}$}
\begin{thebibliography}{\uppercase{THHLB13}}

\bibitem[ASY{\etalchar{*}}19]{Akiba2019Optuna}
\textsc{Akiba T., Sano S., Yanase T., Ohta T., Koyama M.}:
\newblock Optuna: A next-generation hyperparameter optimization framework.
\newblock In \emph{Proceedings of the 25th ACM SIGKDD International Conference
  on Knowledge Discovery and Data Mining} (2019), KDD~'19, ACM, pp.~2623--2631.
\newblock \href {https://doi.org/10.1145/3292500.3330701}
  {\path{doi:10.1145/3292500.3330701}}.

\bibitem[Aut]{automl}
{AutoML} --- {G}oogle {C}loud {AutoML}.
\newblock Accessed February 26, 2021.
\newblock URL: \url{https://cloud.google.com/automl/}.

\bibitem[Bay]{BayesOpt}
{BayesOpt} --- {B}ayesian optimization for science and engineering.
\newblock Accessed February 26, 2021.
\newblock URL: \url{https://bayesopt.github.io/}.

\bibitem[BB12]{Bergstra2012Random}
\textsc{Bergstra J., Bengio Y.}:
\newblock Random search for hyper-parameter optimization.
\newblock \emph{Journal of Machine Learning Research 13} (Feb. 2012), 281--305.
\newblock \href {https://doi.org/10.5555/2188385.2188395}
  {\path{doi:10.5555/2188385.2188395}}.

\bibitem[BBBK11]{Bergstra2011Algorithms}
\textsc{Bergstra J., Bardenet R., Bengio Y., K\'{e}gl B.}:
\newblock Algorithms for hyper-parameter optimization.
\newblock In \emph{Proceedings of the 24th International Conference on Neural
  Information Processing Systems} (2011), NIPS~'11, Curran Associates Inc.,
  pp.~2546--2554.

\bibitem[BBKS13]{Bardenet2013Collaborative}
\textsc{Bardenet R., Brendel M., K\'{e}gl B., Sebag M.}:
\newblock Collaborative hyperparameter tuning.
\newblock In \emph{Proceedings of Machine Learning Research} (2013),
  vol.~28(2), PMLR, pp.~199--207.

\bibitem[BCA{\etalchar{*}}19]{Belkina2019Automated}
\textsc{Belkina A.~C., Ciccolella C.~O., Anno R., Halpert R., Spidlen J.,
  Snyder-Cappione J.~E.}:
\newblock Automated optimized parameters for t-distributed stochastic neighbor
  embedding improve visualization and analysis of large datasets.
\newblock \emph{Nature Communications 10}, 5415 (2019).
\newblock \href {https://doi.org/10.1038/s41467-019-13055-y}
  {\path{doi:10.1038/s41467-019-13055-y}}.

\bibitem[BDA13]{Bekkar2013Evaluation}
\textsc{Bekkar M., Djemaa H.~K., Alitouche T.~A.}:
\newblock Evaluation measures for models assessment over imbalanced data sets.
\newblock \emph{Journal of Information Engineering and Applications 3}, 10
  (2013).

\bibitem[BKE{\etalchar{*}}15]{Bergstra2015Hyperopt}
\textsc{Bergstra J., Komer B., Eliasmith C., Yamins D., Cox D.~D.}:
\newblock Hyperopt: A {Python} library for model selection and hyperparameter
  optimization.
\newblock \emph{Computational Science \& Discovery 8}, 1 (July 2015), 014008.
\newblock \href {https://doi.org/10.1088/1749-4699/8/1/014008}
  {\path{doi:10.1088/1749-4699/8/1/014008}}.

\bibitem[Bre01]{Breiman2001Random}
\textsc{Breiman L.}:
\newblock Random forests.
\newblock \emph{Machine Learning 45} (Oct. 2001), 5--32.
\newblock \href {https://doi.org/10.1023/A:1010933404324}
  {\path{doi:10.1023/A:1010933404324}}.

\bibitem[CAA{\etalchar{*}}19]{Chen2019LDA}
\textsc{Chen S., Andrienko N., Andrienko G., Adilova L., Barlet J., Kindermann
  J., Nguyen P.~H., Thonnard O., Turkay C.}:
\newblock {LDA} ensembles for interactive exploration and categorization of
  behaviors.
\newblock \emph{IEEE Transactions on Visualization and Computer Graphics}
  (2019).
\newblock \href {https://doi.org/10.1109/TVCG.2019.2904069}
  {\path{doi:10.1109/TVCG.2019.2904069}}.

\bibitem[CDM15]{Claesen2015Hyperparameter}
\textsc{Claesen M., De~Moor B.}:
\newblock Hyperparameter search in machine learning.
\newblock In \emph{Proceedings of the 11th Metaheuristics International
  Conference} (2015), MIC~'15.

\bibitem[CG16]{Chen2016XGBoost}
\textsc{Chen T., Guestrin C.}:
\newblock {XGBoost}: A scalable tree boosting system.
\newblock In \emph{Proceedings of the 22nd ACM SIGKDD International Conference
  on Knowledge Discovery and Data Mining} (2016), KDD~'16, ACM, pp.~785--794.
\newblock \href {https://doi.org/10.1145/2939672.2939785}
  {\path{doi:10.1145/2939672.2939785}}.

\bibitem[CGW13]{Cai2013A}
\textsc{Cai J., Garner J.~L., Walkling R.~A.}:
\newblock A paper tiger? {A}n empirical analysis of majority voting.
\newblock \emph{Journal of Corporate Finance 21} (2013), 119--135.
\newblock \href {https://doi.org/10.1016/j.jcorpfin.2013.01.002}
  {\path{doi:10.1016/j.jcorpfin.2013.01.002}}.

\bibitem[CJ20]{Chicco2020The}
\textsc{Chicco D., Jurman G.}:
\newblock The advantages of the {M}atthews correlation coefficient ({MCC}) over
  {F1} score and accuracy in binary classification evaluation.
\newblock \emph{BMC Genomics 21} (Jan. 2020), 6.
\newblock \href {https://doi.org/10.1186/s12864-019-6413-7}
  {\path{doi:10.1186/s12864-019-6413-7}}.

\bibitem[CK05]{Cantu2005An}
\textsc{{Cantu-Paz} E., {Kamath} C.}:
\newblock An empirical comparison of combinations of evolutionary algorithms
  and neural networks for classification problems.
\newblock \emph{IEEE Transactions on Systems, Man, and Cybernetics, Part B
  (Cybernetics) 35}, 5 (Oct. 2005), 915--927.
\newblock \href {https://doi.org/10.1109/TSMCB.2005.847740}
  {\path{doi:10.1109/TSMCB.2005.847740}}.

\bibitem[CMJ{\etalchar{*}}20]{Chatzimparmpas2020The}
\textsc{Chatzimparmpas A., Martins R.~M., Jusufi I., Kucher K., Rossi F.,
  Kerren A.}:
\newblock The state of the art in enhancing trust in machine learning models
  with the use of visualizations.
\newblock \emph{Computer Graphics Forum 39}, 3 (June 2020), 713--756.
\newblock \href {https://doi.org/10.1111/cgf.14034}
  {\path{doi:10.1111/cgf.14034}}.

\bibitem[CMJK20]{Chatzimparmpas2020A}
\textsc{Chatzimparmpas A., Martins R.~M., Jusufi I., Kerren A.}:
\newblock A survey of surveys on the use of visualization for interpreting
  machine learning models.
\newblock \emph{Information Visualization 19}, 3 (July 2020), 207--233.
\newblock \href {https://doi.org/10.1177/1473871620904671}
  {\path{doi:10.1177/1473871620904671}}.

\bibitem[CMK20]{Chatzimparmpas2020t}
\textsc{Chatzimparmpas A., Martins R.~M., Kerren A.}:
\newblock {t-viSNE}: Interactive assessment and interpretation of {t-SNE}
  projections.
\newblock \emph{IEEE Transactions on Visualization and Computer Graphics 26}, 8
  (Aug. 2020), 2696--2714.
\newblock \href {https://doi.org/10.1109/TVCG.2020.2986996}
  {\path{doi:10.1109/TVCG.2020.2986996}}.

\bibitem[CMKK21]{Chatzimparmpas2021StackGenVis}
\textsc{Chatzimparmpas A., Martins R.~M., Kucher K., Kerren A.}:
\newblock {StackGenVis}: Alignment of data, algorithms, and models for stacking
  ensemble learning using performance metrics.
\newblock \emph{IEEE Transactions on Visualization and Computer Graphics}
  (2021).
\newblock \href {https://doi.org/10.1109/TVCG.2020.3030352}
  {\path{doi:10.1109/TVCG.2020.3030352}}.

\bibitem[Com]{comet}
{Comet.ML} --- {B}uild better models faster.
\newblock Accessed February 26, 2021.
\newblock URL: \url{https://comet.ml/}.

\bibitem[CSP{\etalchar{*}}14]{Claesen2014Easy}
\textsc{Claesen M., Simm J., Popovic D., Moreau Y., De~Moor B.}:
\newblock Easy hyperparameter search using {Optunity}.
\newblock \emph{ArXiv e-prints} (Dec. 2014).
\newblock \href {http://arxiv.org/abs/1412.1114} {\path{arXiv:1412.1114}}.

\bibitem[D311]{D3}
{D3} --- {D}ata-driven documents, 2011.
\newblock Accessed February 26, 2021.
\newblock URL: \url{https://d3js.org/}.

\bibitem[Dat]{datarobot}
{DataRobot} --- {E}mpowering the human heroes of the intelligence revolution.
\newblock Accessed February 26, 2021.
\newblock URL: \url{https://www.datarobot.com/}.

\bibitem[DCCE19]{Das2019BEAMES}
\textsc{Das S., Cashman D., Chang R., Endert A.}:
\newblock {BEAMES}: Interactive multi-model steering, selection, and inspection
  for regression tasks.
\newblock \emph{IEEE Computer Graphics and Applications 39}, 9 (Sept. 2019).
\newblock \href {https://doi.org/10.1109/MCG.2019.2922592}
  {\path{doi:10.1109/MCG.2019.2922592}}.

\bibitem[DDF{\etalchar{*}}18]{Francescomarino2018Genetic}
\textsc{{Di Francescomarino} C., Dumas M., Federici M., Ghidini C., Maggi
  F.~M., Rizzi W., Simonetto L.}:
\newblock Genetic algorithms for hyperparameter optimization in predictive
  business process monitoring.
\newblock \emph{Information Systems 74}, Part 1 (May 2018), 67--83.
\newblock \href {https://doi.org/10.1016/j.is.2018.01.003}
  {\path{doi:10.1016/j.is.2018.01.003}}.

\bibitem[DG06]{Davis2006The}
\textsc{Davis J., Goadrich M.}:
\newblock The relationship between precision-recall and {ROC} curves.
\newblock In \emph{Proceedings of the 23rd International Conference on Machine
  Learning} (2006), ICML~'06, ACM, pp.~233--240.
\newblock \href {https://doi.org/10.1145/1143844.1143874}
  {\path{doi:10.1145/1143844.1143874}}.

\bibitem[DG17]{Dua2017}
\textsc{Dua D., Graff C.}:
\newblock {UCI} machine learning repository, 2017.
\newblock Accessed February 26, 2021.
\newblock URL: \url{http://archive.ics.uci.edu/ml}.

\bibitem[FBO{\etalchar{*}}07]{Fiszelew2007Finding}
\textsc{Fiszelew A., Britos P., Ochoa A., Merlino H., Fern{\'a}ndez E.,
  Garc{\'\i}a-Mart{\'\i}nez R.}:
\newblock Finding optimal neural network architecture using genetic algorithms.
\newblock \emph{Research in Computing Science 27} (2007), 15--24.

\bibitem[FH19]{Feurer2019Hyperparameter}
\textsc{Feurer M., Hutter F.}:
\newblock Hyperparameter optimization.
\newblock In \emph{Automated Machine Learning: Methods, Systems, Challenges}.
  Springer International Publishing, 2019, pp.~3--33.
\newblock \href {https://doi.org/10.1007/978-3-030-05318-5_1}
  {\path{doi:10.1007/978-3-030-05318-5_1}}.

\bibitem[FHOM09]{Ferri2009An}
\textsc{Ferri C., Hern{\'a}ndez-Orallo J., Modroiu R.}:
\newblock An experimental comparison of performance measures for
  classification.
\newblock \emph{Pattern Recognition Letters 30}, 1 (Jan. 2009), 27--38.
\newblock \href {https://doi.org/10.1016/j.patrec.2008.08.010}
  {\path{doi:10.1016/j.patrec.2008.08.010}}.

\bibitem[FKH18]{Falkner2018BOHB}
\textsc{Falkner S., Klein A., Hutter F.}:
\newblock {BOHB}: Robust and efficient hyperparameter optimization at scale.
\newblock In \emph{Proceedings of Machine Learning Research} (2018), vol.~80,
  PMLR, pp.~1437--1446.

\bibitem[Fla10]{Flask}
{Flask} --- {A} micro web framework written in {Python}, 2010.
\newblock Accessed February 26, 2021.
\newblock URL: \url{https://flask.palletsprojects.com/}.

\bibitem[FSA99]{Freund1999A}
\textsc{Freund Y., Schapire R., Abe N.}:
\newblock A short introduction to boosting.
\newblock \emph{Journal of Japanese Society for Artificial Intelligence 14}, 5
  (Sept. 1999), 771--780.

\bibitem[GSM{\etalchar{*}}17]{Golovin2017Google}
\textsc{Golovin D., Solnik B., Moitra S., Kochanski G., Karro J., Sculley D.}:
\newblock {Google} {Vizier}: A service for black-box optimization.
\newblock In \emph{Proceedings of the 23rd ACM SIGKDD International Conference
  on Knowledge Discovery and Data Mining} (2017), KDD~'17, ACM, pp.~1487--1495.
\newblock \href {https://doi.org/10.1145/3097983.3098043}
  {\path{doi:10.1145/3097983.3098043}}.

\bibitem[HDK{\etalchar{*}}19]{Hamid2019Visual}
\textsc{Hamid S., Derstroff A., Klemm S., Ngo Q.~Q., Jiang X., Linsen L.}:
\newblock Visual ensemble analysis to study the influence of hyper-parameters
  on training deep neural networks.
\newblock In \emph{Proceedings of the EuroVis Workshop on Machine Learning
  Methods in Visualisation for Big Data} (2019), MLVis~'19, The Eurographics
  Association.
\newblock \href {https://doi.org/10.2312/mlvis.20191160}
  {\path{doi:10.2312/mlvis.20191160}}.

\bibitem[HHLB11]{Hutter2011Sequential}
\textsc{Hutter F., Hoos H.~H., Leyton-Brown K.}:
\newblock Sequential model-based optimization for general algorithm
  configuration.
\newblock In \emph{Proceedings of the International Conference on Learning and
  Intelligent Optimization} (2011), LION~'11, Springer Berlin Heidelberg,
  pp.~507--523.
\newblock \href {https://doi.org/10.1007/978-3-642-25566-3_40}
  {\path{doi:10.1007/978-3-642-25566-3_40}}.

\bibitem[HHLB13]{Hutter2013An}
\textsc{Hutter F., Hoos H.~H., Leyton-Brown K.}:
\newblock An efficient approach for assessing parameter importance in bayesian
  optimization.
\newblock In \emph{Proceedings of the NIPS workshop on Bayesian Optimization in
  Theory and Practice} (2013), BayesOpt~'13.

\bibitem[HHLB14]{Hutter2014An}
\textsc{Hutter F., Hoos H., Leyton-Brown K.}:
\newblock An efficient approach for assessing hyperparameter importance.
\newblock In \emph{Proceedings of Machine Learning Research} (2014),
  vol.~32(1), PMLR, pp.~754--762.

\bibitem[HHLBS09]{Hutter2009ParamILS}
\textsc{Hutter F., Hoos H.~H., Leyton-Brown K., St\"{u}tzle T.}:
\newblock {ParamILS}: An automatic algorithm configuration framework.
\newblock \emph{Journal of Artificial Intelligence Research 36}, 1 (Oct. 2009),
  267--306.
\newblock \href {https://doi.org/10.1613/jair.2861}
  {\path{doi:10.1613/jair.2861}}.

\bibitem[ID87]{Inselberg1987Parallel}
\textsc{Inselberg A., Dimsdale B.}:
\newblock Parallel coordinates for visualizing multi-dimensional geometry.
\newblock In \emph{Proceedings of the 5th International Conference on Computer
  Graphics} (1987), CG~'87, Springer Japan, pp.~25--44.
\newblock \href {https://doi.org/10.1007/978-4-431-68057-4_3}
  {\path{doi:10.1007/978-4-431-68057-4_3}}.

\bibitem[JDO{\etalchar{*}}17]{Jaderberg2017Population}
\textsc{Jaderberg M., Dalibard V., Osindero S., Czarnecki W.~M., Donahue J.,
  Razavi A., Vinyals O., Green T., Dunning I., Simonyan K., Fernando C.,
  Kavukcuoglu K.}:
\newblock Population based training of neural networks.
\newblock \emph{ArXiv e-prints} (Nov. 2017).
\newblock \href {http://arxiv.org/abs/1711.09846} {\path{arXiv:1711.09846}}.

\bibitem[JES{\etalchar{*}}20]{Jonsson2020Visual}
\textsc{J{\"o}nsson D., Eilertsen G., Shi H., Zheng J., Ynnerman A., Unger J.}:
\newblock Visual analysis of the impact of neural network hyper-parameters.
\newblock In \emph{Proceedings of the EGEV International Workshop on Machine
  Learning Methods in Visualisation for Big Data} (2020), MLVis~'20, The
  Eurographics Association.
\newblock \href {https://doi.org/10.2312/mlvis.20201101}
  {\path{doi:10.2312/mlvis.20201101}}.

\bibitem[JWXY16]{Jia2016QIM}
\textsc{Jia D., Wang R., Xu C., Yu Z.}:
\newblock {QIM}: Quantifying hyperparameter importance for deep learning.
\newblock In \emph{Proceedings of the IFIP International Conference on Network
  and Parallel Computing} (2016), NPC~16, Springer International Publishing,
  pp.~180--188.
\newblock \href {https://doi.org/10.1007/978-3-319-47099-3_15}
  {\path{doi:10.1007/978-3-319-47099-3_15}}.

\bibitem[KB19]{Kobak2019The}
\textsc{Kobak D., Berens P.}:
\newblock The art of using {t-SNE} for single-cell transcriptomics.
\newblock \emph{Nature Communications 10}, 5416 (Nov. 2019).
\newblock \href {https://doi.org/10.1038/s41467-019-13056-x}
  {\path{doi:10.1038/s41467-019-13056-x}}.

\bibitem[KE05]{Kerren2005EAVis}
\textsc{Kerren A., Egger T.}:
\newblock {EAVis}: A visualization tool for evolutionary algorithms.
\newblock In \emph{Proceedings of the IEEE Symposium on Visual Languages and
  Human-Centric Computing} (2005), VL/HCC~'05, IEEE, pp.~299--301.
\newblock \href {https://doi.org/10.1109/VLHCC.2005.33}
  {\path{doi:10.1109/VLHCC.2005.33}}.

\bibitem[Ker06]{Kerren2006Improving}
\textsc{Kerren A.}:
\newblock Improving strategy parameters of evolutionary computations with
  interactive coordinated views.
\newblock In \emph{Proceedings of the IASTED International Conference on
  Visualization, Imaging, and Image Processing} (2006), VIIP~'06, ACTA Press,
  pp.~88--93.

\bibitem[KGG{\etalchar{*}}18]{Koch2018Autotune}
\textsc{Koch P., Golovidov O., Gardner S., Wujek B., Griffin J., Xu Y.}:
\newblock Autotune: A derivative-free optimization framework for hyperparameter
  tuning.
\newblock In \emph{Proceedings of the 24th ACM SIGKDD International Conference
  on Knowledge Discovery and Data Mining} (2018), KDD~'18, ACM, pp.~443--452.
\newblock \href {https://doi.org/10.1145/3219819.3219837}
  {\path{doi:10.1145/3219819.3219837}}.

\bibitem[KKP{\etalchar{*}}18]{Jingwoong2018CHOPT}
\textsc{Kim J., Kim M., Park H., Kusdavletov E., Lee D., Kim A., Kim J.-H., Ha
  J.-W., Sung N.}:
\newblock {CHOPT}: Automated hyperparameter optimization framework for
  cloud-based machine learning platforms.
\newblock \emph{ArXiv e-prints} (Oct. 2018).
\newblock \href {http://arxiv.org/abs/1810.03527} {\path{arXiv:1810.03527}}.

\bibitem[KMF{\etalchar{*}}17]{Ke2017LightGBM}
\textsc{Ke G., Meng Q., Finley T., Wang T., Chen W., Ma W., Ye Q., Liu T.-Y.}:
\newblock {LightGBM}: A highly efficient gradient boosting decision tree.
\newblock In \emph{Proceedings of the 31st International Conference on Neural
  Information Processing Systems} (2017), NIPS~'17, Curran Associates Inc.,
  pp.~3149--3157.

\bibitem[Kru64]{Kruskal1964Multidimensional}
\textsc{Kruskal J.~B.}:
\newblock Multidimensional scaling by optimizing goodness of fit to a nonmetric
  hypothesis.
\newblock \emph{Psychometrika 29}, 1 (Mar. 1964), 1--27.
\newblock \href {https://doi.org/10.1007/BF02289565}
  {\path{doi:10.1007/BF02289565}}.

\bibitem[LCW{\etalchar{*}}18]{Li2018HyperTuner}
\textsc{Li T., Convertino G., Wang W., Most H., Zajonc T., Tsai Y.}:
\newblock {HyperTuner}: Visual analytics for hyperparameter tuning by
  professionals.
\newblock In \emph{Proceedings of the IEEE VIS Workshop on Machine Learning
  from User Interaction for Visualization and Analytics} (2018), MLUI~'18.

\bibitem[LH18]{Liu2018Somewhere}
\textsc{Liu Y., Heer J.}:
\newblock Somewhere over the rainbow: An empirical assessment of quantitative
  colormaps.
\newblock In \emph{Proceedings of the 2018 CHI Conference on Human Factors in
  Computing Systems} (2018), CHI~'18, ACM, pp.~598:1--598:12.
\newblock \href {https://doi.org/10.1145/3173574.3174172}
  {\path{doi:10.1145/3173574.3174172}}.

\bibitem[LJ19]{Latha2019Improving}
\textsc{Latha C. B.~C., Jeeva S.~C.}:
\newblock Improving the accuracy of prediction of heart disease risk based on
  ensemble classification techniques.
\newblock \emph{Informatics in Medicine Unlocked 16} (2019), 100203.
\newblock \href {https://doi.org/10.1016/j.imu.2019.100203}
  {\path{doi:10.1016/j.imu.2019.100203}}.

\bibitem[LJD{\etalchar{*}}17]{Li2017Hyperband}
\textsc{Li L., Jamieson K., DeSalvo G., Rostamizadeh A., Talwalkar A.}:
\newblock Hyperband: A novel bandit-based approach to hyperparameter
  optimization.
\newblock \emph{Jounal of Machine Learning Research 18}, 1 (Jan. 2017),
  6765--6816.

\bibitem[LJVR08]{Lobo2008AUC}
\textsc{Lobo J.~M., Jim{\'e}nez-Valverde A., Real R.}:
\newblock {AUC}: A misleading measure of the performance of predictive
  distribution models.
\newblock \emph{Global Ecology and Biogeography 17}, 2 (Mar. 2008), 145--151.
\newblock \href {https://doi.org/10.1111/j.1466-8238.2007.00358.x}
  {\path{doi:10.1111/j.1466-8238.2007.00358.x}}.

\bibitem[LLN{\etalchar{*}}18]{Liaw2018Tune}
\textsc{Liaw R., Liang E., Nishihara R., Moritz P., Gonzalez J.~E., Stoica I.}:
\newblock Tune: A research platform for distributed model selection and
  training.
\newblock In \emph{Proceedings of the ICML/IJCAI-ECAI International Workshop on
  Automatic Machine Learning} (2018), AutoML~'18.

\bibitem[LTKS19]{Liu2019Auptimizer}
\textsc{{Liu} J., {Tripathi} S., {Kurup} U., {Shah} M.}:
\newblock Auptimizer --- {A}n extensible, open-source framework for
  hyperparameter tuning.
\newblock In \emph{Proceedings of the IEEE International Conference on Big
  Data} (2019), Big Data~'19, IEEE, pp.~339--348.
\newblock \href {https://doi.org/10.1109/BigData47090.2019.9006330}
  {\path{doi:10.1109/BigData47090.2019.9006330}}.

\bibitem[LXL{\etalchar{*}}18]{Liu2018Visual}
\textsc{Liu S., Xiao J., Liu J., Wang X., Wu J., Zhu J.}:
\newblock Visual diagnosis of tree boosting methods.
\newblock \emph{IEEE Transactions on Visualization and Computer Graphics 24}, 1
  (Jan. 2018), 163--173.
\newblock \href {https://doi.org/10.1109/TVCG.2017.2744378}
  {\path{doi:10.1109/TVCG.2017.2744378}}.

\bibitem[MHM18]{McInnes2018UMAP}
\textsc{{McInnes} L., {Healy} J., {Melville} J.}:
\newblock {UMAP}: Uniform manifold approximation and projection for dimension
  reduction.
\newblock \emph{ArXiv e-prints 1802.03426} (Feb. 2018).
\newblock \href {http://arxiv.org/abs/1802.03426} {\path{arXiv:1802.03426}}.

\bibitem[MRB{\etalchar{*}}13]{Mansouri2013Quantitative}
\textsc{Mansouri K., Ringsted T., Ballabio D., Todeschini R., Consonni V.}:
\newblock Quantitative structure–activity relationship models for ready
  biodegradability of chemicals.
\newblock \emph{Journal of Chemical Information and Modeling 53}, 4 (2013),
  867--878.
\newblock \href {https://doi.org/10.1021/ci4000213}
  {\path{doi:10.1021/ci4000213}}.

\bibitem[MRK06]{McNee2006Being}
\textsc{McNee S.~M., Riedl J., Konstan J.~A.}:
\newblock Being accurate is not enough: How accuracy metrics have hurt
  recommender systems.
\newblock In \emph{CHI~'06 Extended Abstracts on Human Factors in Computing
  Systems} (2006), CHI EA~'06, ACM, pp.~1097--1101.
\newblock \href {https://doi.org/10.1145/1125451.1125659}
  {\path{doi:10.1145/1125451.1125659}}.

\bibitem[MXLM20]{Ma2020Explaining}
\textsc{Ma Y., Xie T., Li J., Maciejewski R.}:
\newblock Explaining vulnerabilities to adversarial machine learning through
  visual analytics.
\newblock \emph{IEEE Transactions on Visualization and Computer Graphics 26}, 1
  (Jan. 2020), 1075--1085.
\newblock \href {https://doi.org/10.1109/TVCG.2019.2934631}
  {\path{doi:10.1109/TVCG.2019.2934631}}.

\bibitem[NNI]{nni}
{NNI} --- {Microsoft} {N}eural {N}etwork {I}ntelligence.
\newblock Accessed February 26, 2021.
\newblock URL: \url{https://github.com/microsoft/nni}.

\bibitem[NP20]{Neto2020Explainable}
\textsc{Neto M.~P., Paulovich F.~V.}:
\newblock {Explainable} {Matrix} --- {V}isualization for global and local
  interpretability of random forest classification ensembles.
\newblock \emph{IEEE Transactions on Visualization and Computer Graphics}
  (2020).
\newblock \href {https://doi.org/10.1109/TVCG.2020.3030354}
  {\path{doi:10.1109/TVCG.2020.3030354}}.

\bibitem[PBB19]{Probst2019Tunability}
\textsc{Probst P., Boulesteix A.-L., Bischl B.}:
\newblock Tunability: Importance of hyperparameters of machine learning
  algorithms.
\newblock \emph{Journal of Machine Learning Research 20}, 53 (2019), 1--32.

\bibitem[PKK{\etalchar{*}}19]{Park2019VisualHyperTuner}
\textsc{Park H., Kim J., Kim M., Kim J., Choo J., Ha J., Sung N.}:
\newblock {VisualHyperTuner}: Visual analytics for user-driven hyperparameter
  tuning of deep neural networks.
\newblock In \emph{Proceedings of the 2nd SysML Conference} (2019), SysML~19.

\bibitem[plo10]{plotly}
{Plotly} --- {JavaScript} open source graphing library, 2010.
\newblock Accessed February 26, 2021.
\newblock URL: \url{https://plotly.com}.

\bibitem[PN17]{Pereira2017A}
\textsc{Pereira L., Nunes N.}:
\newblock A comparison of performance metrics for event classification in
  non-intrusive load monitoring.
\newblock In \emph{Proceedings of the IEEE International Conference on Smart
  Grid Communications} (2017), SmartGridComm~'17, IEEE, pp.~159--164.
\newblock \href {https://doi.org/10.1109/SmartGridComm.2017.8340682}
  {\path{doi:10.1109/SmartGridComm.2017.8340682}}.

\bibitem[PNKC21]{Park2021HyperTendril}
\textsc{Park H., Nam Y., Kim J., Choo J.}:
\newblock {HyperTendril}: Visual analytics for user-driven hyperparameter
  optimization of deep neural networks.
\newblock \emph{IEEE Transactions on Visualization \& Computer Graphics}
  (2021).
\newblock \href {https://doi.org/10.1109/TVCG.2020.3030380}
  {\path{doi:10.1109/TVCG.2020.3030380}}.

\bibitem[Pow11]{Powers2011Evaluation}
\textsc{Powers D. M.~W.}:
\newblock Evaluation: From precision, recall and {F}-measure to {ROC},
  informedness, markedness \& correlation.
\newblock \emph{Journal of Machine Learning Technologies 2}, 1 (2011), 37--63.

\bibitem[PVG{\etalchar{*}}11]{Pedregosa2011Scikit}
\textsc{Pedregosa F., Varoquaux G., Gramfort A., Michel V., Thirion B., Grisel
  O., Blondel M., Prettenhofer P., Weiss R., Dubourg V., Vanderplas J., Passos
  A., Cournapeau D., Brucher M., Perrot M., Duchesnay E.}:
\newblock {Scikit-Learn}: Machine learning in {P}ython.
\newblock \emph{Journal of Machine Learning Research 12} (Nov. 2011),
  2825--2830.
\newblock \href {https://doi.org/10.5555/1953048.2078195}
  {\path{doi:10.5555/1953048.2078195}}.

\bibitem[SDC{\etalchar{*}}17]{Swearingen2017ATM}
\textsc{{Swearingen} T., {Drevo} W., {Cyphers} B., {Cuesta-Infante} A., {Ross}
  A., {Veeramachaneni} K.}:
\newblock {ATM}: A distributed, collaborative, scalable system for automated
  machine learning.
\newblock In \emph{Proceedings of the IEEE International Conference on Big
  Data} (2017), Big Data~'17, IEEE, pp.~151--162.
\newblock \href {https://doi.org/10.1109/BigData.2017.8257923}
  {\path{doi:10.1109/BigData.2017.8257923}}.

\bibitem[SJS{\etalchar{*}}18]{Schneider2018Integrating}
\textsc{Schneider B., J{\"{a}}ckle D., Stoffel F., Diehl A., Fuchs J., Keim
  D.~A.}:
\newblock Integrating data and model space in ensemble learning by visual
  analytics.
\newblock \emph{IEEE Transactions on Big Data} (2018).
\newblock \href {https://doi.org/10.1109/TBDATA.2018.2877350}
  {\path{doi:10.1109/TBDATA.2018.2877350}}.

\bibitem[SKJ{\etalchar{*}}17]{Sung2017NSML}
\textsc{Sung N., Kim M., Jo H., Yang Y., Kim J., Lausen L., Kim Y., Lee G.,
  Kwak D., Ha J.-W., Kim S.}:
\newblock {NSML}: A machine learning platform that enables you to focus on your
  models.
\newblock In \emph{Proceedings of the NIPS Workshop on ML Systems} (2017),
  ML-Sys~'17.

\bibitem[SL09]{Sokolova2009Performance}
\textsc{Sokolova M., Lapalme G.}:
\newblock A systematic analysis of performance measures for classification
  tasks.
\newblock \emph{Information Processing \& Management 45}, 4 (July 2009),
  427--437.
\newblock \href {https://doi.org/10.1016/j.ipm.2009.03.002}
  {\path{doi:10.1016/j.ipm.2009.03.002}}.

\bibitem[SLA12]{Snoek2012Practical}
\textsc{Snoek J., Larochelle H., Adams R.~P.}:
\newblock Practical {B}ayesian optimization of machine learning algorithms.
\newblock In \emph{Proceedings of the 25th International Conference on Neural
  Information Processing Systems} (2012), vol.~2 of \emph{NIPS~'12}, Curran
  Associates Inc., pp.~2951--2959.

\bibitem[SPG14]{Stolper2014Progressive}
\textsc{Stolper C.~D., Perer A., Gotz D.}:
\newblock Progressive visual analytics: User-driven visual exploration of
  in-progress analytics.
\newblock \emph{IEEE Transactions on Visualization and Computer Graphics 20},
  12 (Dec. 2014), 1653--1662.
\newblock \href {https://doi.org/10.1109/TVCG.2014.2346574}
  {\path{doi:10.1109/TVCG.2014.2346574}}.

\bibitem[SR15]{Saito2015The}
\textsc{Saito T., Rehmsmeier M.}:
\newblock The precision-recall plot is more informative than the {ROC} plot
  when evaluating binary classifiers on imbalanced datasets.
\newblock \emph{PLOS ONE 10}, 3 (Mar. 2015), e0118432.
\newblock \href {https://doi.org/10.1371/journal.pone.0118432}
  {\path{doi:10.1371/journal.pone.0118432}}.

\bibitem[SR18]{Sagi2018Ensemble}
\textsc{Sagi O., Rokach L.}:
\newblock Ensemble learning: A survey.
\newblock \emph{WIREs Data Mining and Knowledge Discovery 8}, 4 (July--Aug.
  2018), e1249.
\newblock \href {https://doi.org/10.1002/widm.1249}
  {\path{doi:10.1002/widm.1249}}.

\bibitem[SSW{\etalchar{*}}16]{Shahriari2016Taking}
\textsc{{Shahriari} B., {Swersky} K., {Wang} Z., {Adams} R.~P., {de Freitas}
  N.}:
\newblock Taking the human out of the loop: A review of {B}ayesian
  optimization.
\newblock \emph{Proceedings of the IEEE 104}, 1 (2016), 148--175.
\newblock \href {https://doi.org/10.1109/JPROC.2015.2494218}
  {\path{doi:10.1109/JPROC.2015.2494218}}.

\bibitem[Stu13]{Sturm2013Classification}
\textsc{Sturm B.~L.}:
\newblock Classification accuracy is not enough.
\newblock \emph{Journal of Intelligent Information Systems 41}, 3 (Dec. 2013),
  371--406.
\newblock \href {https://doi.org/10.1007/s10844-013-0250-y}
  {\path{doi:10.1007/s10844-013-0250-y}}.

\bibitem[Tak01]{Takagi2001Interactive}
\textsc{Takagi H.}:
\newblock Interactive evolutionary computation: Fusion of the capabilities of
  {EC} optimization and human evaluation.
\newblock \emph{Proceedings of the IEEE 89}, 9 (Sept. 2001), 1275--1296.
\newblock \href {https://doi.org/10.1109/5.949485}
  {\path{doi:10.1109/5.949485}}.

\bibitem[TBCT{\etalchar{*}}18]{Tsirigotis2018Orion}
\textsc{Tsirigotis C., Bouthillier X., Corneau-Tremblay F., Henderson P.,
  Askari R., Lavoie-Marchildon S., Deleu T., Suhubdy D., Noukhovitch M.,
  Bastien F., Lamblin P.}:
\newblock Or{\'i}on: Experiment version control for efficient hyperparameter
  optimization.
\newblock In \emph{Proceedings of the ICML Workshop on Reproducibility in
  Machine Learning} (2018), RML~'18.

\bibitem[Tha18]{Tharwat2018Classification}
\textsc{Tharwat A.}:
\newblock Classification assessment methods.
\newblock \emph{Applied Computing and Informatics} (2018).
\newblock \href {https://doi.org/10.1016/j.aci.2018.08.003}
  {\path{doi:10.1016/j.aci.2018.08.003}}.

\bibitem[THHLB13]{Thornton2013Auto}
\textsc{Thornton C., Hutter F., Hoos H.~H., Leyton-Brown K.}:
\newblock {Auto-WEKA}: Combined selection and hyperparameter optimization of
  classification algorithms.
\newblock In \emph{Proceedings of the 19th ACM SIGKDD International Conference
  on Knowledge Discovery and Data Mining} (2013), KDD~'13, ACM, pp.~847--855.
\newblock \href {https://doi.org/10.1145/2487575.2487629}
  {\path{doi:10.1145/2487575.2487629}}.

\bibitem[TLKT09]{Talbot2009EnsembleMatrix}
\textsc{Talbot J., Lee B., Kapoor A., Tan D.~S.}:
\newblock {EnsembleMatrix}: Interactive visualization to support machine
  learning with multiple classifiers.
\newblock In \emph{Proceedings of the SIGCHI Conference on Human Factors in
  Computing Systems} (2009), CHI~'09, ACM, pp.~1283--1292.
\newblock \href {https://doi.org/10.1145/1518701.1518895}
  {\path{doi:10.1145/1518701.1518895}}.

\bibitem[TMB{\etalchar{*}}18]{Tsay2018Runway}
\textsc{Tsay J., Mummert T., Bobroff N., Braz A., Westerink P., Hirzel M.,
  Heights Y.}:
\newblock Runway: Machine learning model experiment management tool.
\newblock In \emph{Proceedings of the 1st SysML Conference} (2018), SysML~'18.

\bibitem[TPB{\etalchar{*}}18]{Turkay2018Progressive}
\textsc{Turkay C., Pezzotti N., Binnig C., Strobelt H., Hammer B., Keim D.~A.,
  Fekete J., Palpanas T., Wang Y., Rusu F.}:
\newblock Progressive data science: Potential and challenges.
\newblock \emph{CoRR abs/1812.08032} (2018).
\newblock \href {http://arxiv.org/abs/1812.08032} {\path{arXiv:1812.08032}}.

\bibitem[vdMH08]{vanDerMaaten2008Visualizing}
\textsc{van~der Maaten L., Hinton G.}:
\newblock Visualizing data using {t-SNE}.
\newblock \emph{Journal of Machine Learning Research 9} (2008), 2579--2605.

\bibitem[vRH17]{Rijn2017An}
\textsc{van Rijn J.~N., Hutter F.}:
\newblock An empirical study of hyperparameter importance across datasets.
\newblock In \emph{Proceedings of the ECML-PKDD International Workshop on
  Automatic Machine Learning} (2017), AutoML~'17.

\bibitem[vRH18]{Rijn2018Hyperparameter}
\textsc{van Rijn J.~N., Hutter F.}:
\newblock Hyperparameter importance across datasets.
\newblock In \emph{Proceedings of the 24th ACM SIGKDD International Conference
  on Knowledge Discovery and Data Mining} (2018), KDD~'18, ACM, pp.~2367--2376.
\newblock \href {https://doi.org/10.1145/3219819.3220058}
  {\path{doi:10.1145/3219819.3220058}}.

\bibitem[vue14]{vuejs}
{Vue.js} --- {T}he progressive {JavaScript} framework, 2014.
\newblock Accessed February 26, 2021.
\newblock URL: \url{https://vuejs.org/}.

\bibitem[WMJ{\etalchar{*}}19]{Wang2019ATMSeer}
\textsc{Wang Q., Ming Y., Jin Z., Shen Q., Liu D., Smith M.~J., Veeramachaneni
  K., Qu H.}:
\newblock {ATMSeer}: Increasing transparency and controllability in automated
  machine learning.
\newblock In \emph{Proceedings of the 2019 CHI Conference on Human Factors in
  Computing Systems} (2019), CHI~'19, ACM, pp.~681:1--681:12.
\newblock \href {https://doi.org/10.1145/3290605.3300911}
  {\path{doi:10.1145/3290605.3300911}}.

\bibitem[Wol92]{Wolpert1992Stacked}
\textsc{Wolpert D.~H.}:
\newblock Stacked generalization.
\newblock \emph{Neural Networks 5}, 2 (1992), 241--259.
\newblock \href {https://doi.org/10.1016/S0893-6080(05)80023-1}
  {\path{doi:10.1016/S0893-6080(05)80023-1}}.

\bibitem[WRW{\etalchar{*}}20]{Wang2020AutoAI}
\textsc{Wang D., Ram P., Weidele D. K.~I., Liu S., Muller M., Weisz J.~D.,
  Valente A., Chaudhary A., Torres D., Samulowitz H., Amini L.}:
\newblock {AutoAI}: Automating the end-to-end ai lifecycle with
  humans-in-the-loop.
\newblock In \emph{Proceedings of the 25th International Conference on
  Intelligent User Interfaces Companion} (2020), IUI~'20, ACM, pp.~77--78.
\newblock \href {https://doi.org/10.1145/3379336.3381474}
  {\path{doi:10.1145/3379336.3381474}}.

\bibitem[WWO{\etalchar{*}}20]{Weidele2020AutoAIViz}
\textsc{Weidele D. K.~I., Weisz J.~D., Oduor E., Muller M., Andres J., Gray A.,
  Wang D.}:
\newblock {AutoAIViz}: Opening the blackbox of automated artificial
  intelligence with conditional parallel coordinates.
\newblock In \emph{Proceedings of the 25th International Conference on
  Intelligent User Interfaces} (2020), IUI~'20, ACM, pp.~308--312.
\newblock \href {https://doi.org/10.1145/3377325.3377538}
  {\path{doi:10.1145/3377325.3377538}}.

\bibitem[XXM{\etalchar{*}}19]{Xu2019EnsembleLens}
\textsc{Xu K., Xia M., Mu X., Wang Y., Cao N.}:
\newblock {EnsembleLens}: Ensemble-based visual exploration of anomaly
  detection algorithms with multidimensional data.
\newblock \emph{IEEE Transactions on Visualization and Computer Graphics 25}, 1
  (Jan. 2019), 109--119.
\newblock \href {https://doi.org/10.1109/TVCG.2018.2864825}
  {\path{doi:10.1109/TVCG.2018.2864825}}.

\bibitem[YM14]{Yogatama2014Efficient}
\textsc{Yogatama D., Mann G.}:
\newblock Efficient transfer learning method for automatic hyperparameter
  tuning.
\newblock In \emph{Proceedings of Machine Learning Research} (Apr. 2014),
  vol.~33, PMLR, pp.~1077--1085.

\bibitem[YRK{\etalchar{*}}15]{Young2015Optimizing}
\textsc{Young S.~R., Rose D.~C., Karnowski T.~P., Lim S.-H., Patton R.~M.}:
\newblock Optimizing deep learning hyper-parameters through an evolutionary
  algorithm.
\newblock In \emph{Proceedings of the Workshop on Machine Learning in
  High-Performance Computing Environments} (2015), MLHPC~'15, ACM.
\newblock \href {https://doi.org/10.1145/2834892.2834896}
  {\path{doi:10.1145/2834892.2834896}}.

\bibitem[ZWLC19]{Zhao2019iForest}
\textsc{Zhao X., Wu Y., Lee D.~L., Cui W.}:
\newblock {iForest}: Interpreting random forests via visual analytics.
\newblock \emph{IEEE Transactions on Visualization and Computer Graphics 25}, 1
  (Jan. 2019), 407--416.
\newblock \href {https://doi.org/10.1109/TVCG.2018.2864475}
  {\path{doi:10.1109/TVCG.2018.2864475}}.

\bibitem[ZWM{\etalchar{*}}19]{Zhang2019Manifold}
\textsc{Zhang J., Wang Y., Molino P., Li L., Ebert D.~S.}:
\newblock {Manifold}: A model-agnostic framework for interpretation and
  diagnosis of machine learning models.
\newblock \emph{IEEE Transactions on Visualization and Computer Graphics 25}, 1
  (Jan. 2019), 364--373.
\newblock \href {https://doi.org/10.1109/TVCG.2018.2864499}
  {\path{doi:10.1109/TVCG.2018.2864499}}.

\end{thebibliography}
